\documentclass[12pt]{article}
\usepackage[left=1in, right=1in, top=1in, bottom=1in]{geometry}
\usepackage{array}
\usepackage{amsmath,amssymb,amsthm}
\usepackage{bm}
\usepackage{graphicx,subfig,psfrag,epsf}
\usepackage{booktabs}
\usepackage{setspace}
\usepackage{natbib}
\usepackage{multirow}

\RequirePackage[OT1]{fontenc}
\RequirePackage{amsthm,amsmath}

\newtheorem{thm}{Theorem}

\usepackage{graphicx}
\usepackage{amssymb}
\usepackage{epsfig, graphicx, amsmath}
\usepackage{graphicx}
\usepackage{natbib}
\usepackage{multirow}

\newcommand{\Mean}{{\mbox{E}}}
\newcommand{\Var}{{\mbox{Var}}}

\newcommand{\diag}{{\mbox{diag}}}
\newcommand{\prob}{{\mbox{Pr}}}
\newtheorem{coro}{Corollary}[section]
\newtheorem{prop}{Proposition}[section]
\newtheorem*{note}{Notation}
\newtheorem{lemma}{Lemma}
\newtheorem{remark}{Remark}[section]
\newtheorem{exam}{Example}
\newtheorem{cond}{Condition}[section]

\begin{document}
	
\title{Robust Learning for Optimal Treatment Decision with NP-Dimensionality}
\author{Chengchun Shi, Rui Song and Wenbin Lu}
\date{}                                           
\maketitle

\begin{abstract}
	In order to identify important variables that are involved in making optimal treatment decision, \cite{lv2011} proposed a penalized least squared regression framework for a fixed number of predictors, which is robust against the misspecification of the conditional mean model. Two problems arise: (i) in a world of explosively big data, effective methods are needed to handle ultra-high dimensional data set, for example, with the dimension of predictors is of the non-polynomial (NP) order of the sample size; (ii) both the propensity score and conditional mean models need to be estimated from data under NP dimensionality.
	
	In this paper, we propose a two-step estimation procedure for deriving the optimal treatment regime under NP dimensionality. In both steps, penalized regressions are employed with the non-concave penalty function, where the conditional mean model of the response given predictors may be misspecified. The asymptotic properties, such as weak oracle properties, selection consistency and oracle distributions, of the proposed estimators are investigated. In addition, we study the limiting distribution of the estimated value function for the obtained optimal treatment regime. The empirical performance of the proposed estimation method is evaluated by simulations and an application to a depression dataset from the STAR*D study.
\end{abstract}

\noindent%
\vfill


\section{Introduction}
Personalized medicine, which has gained much attentions over a few past years, is a medical paradigm that emphasizes systematic use of individual patient information to optimize that patient's health care. In this paradigm, the primary interest lies in identifying the optimal treatment strategy that assigns the best treatment to a patient based on his/her observed covariates.
Formally speaking, a treatment regime is a function that maps the sample space of patient's covariates to the treatments. 

There is a growing literature for estimating the optimal individualized treatment regimes. Existing literature can be casted into as model based methods and direct search methods.
Popular model based methods include $Q$-learning \citep{watkins1992, chak2010} and $A$-learning \citep{robins2000,Murphy2005}, where $Q$-learning models the conditional mean of the response given predictors and treatment while $A$-learning models the contrast function. The advantage of $A$-learning is robust against the misspecification of the baseline mean function, provided that the propensity score model is correctly specified.
Recently, \citet{zhang2012} proposed inverse propensity score weighted (IPSW) and augmented-IPSW estimators to directly maximize the mean potential outcome under a given treatment regime, i.e. the value function. Moreover, \cite{zhao2012} and \cite{zhang2012} recast the estimation of the value function from a classification perspective and use machine
learning tools, to directly search for the optimal treatment regimes.

The rapid advances and breakthrough in technology and communication systems make it possible to gather an extraordinary large number of prognostic factors for each individual. For example, in the Sequenced Treatment Alternative to Relieve Depression (STAR*D) study, over $350$ covariates are collected from each patient. With such data gathered at hand, it is of significant importance to organize and integrate information that is relevant to make optimal individualized treatment decisions, which makes variable selection as an emerging need for implementing personalized medicine. In addition, it is common to encounter studies where the number of covariates is comparable or much larger than the sample size. Therefore, variable selection is also essential in making high-dimensional statistical inference for estimated optimal treatment regimes.  There have been extensive developments of variable selection methods for prediction, for example, LASSO \citep{tibsh1996}, SCAD \citep{fan2001}, MCP \citep{Zhang09} and many others in the context of penalized regression. Their associated inferential properties have been studied when the number of predictors is fixed, diverging with the sample size and of the non-polynomial order of the sample size.



In contrast to the large amount of work on developing variable selection methods for prediction, the variable selection tools for deriving optimal individualized treatment regimes have been less studied, especially when the number of predictors is much larger than the sample size. Among them available, \cite{gunter2011} proposed variable ranking methods for the marginal qualitative interaction of predictors with treatment. \cite{fan2015} developed a sequential advantage selection method that extends the marginal ranking methods by selecting important variables with qualitative interaction in a sequential fashion.  However, no theoretical justifications are provided for these methods.
\cite{qian2011} proposed to estimate the conditional mean response using a $L_1$-penalized regression and studied the error bound of the value function for the estimated treatment regime. However, the associated variable selection properties, such as selection consistency, convergence rate and oracle distribution, are not studied. \cite{lv2011} introduced a new penalized least squared regression framework, which is robust against the misspecification of the conditional mean function. However, they only studied the case when the number of covariates is fixed and the propensity score model is known as in randomized clinical trials. \cite{Sgroup13} proposed penalized outcome weighted learning for the case with the fixed number of predictors.


In this paper, 
we study the penalized least squared regression framework considered in \cite{lv2011} when the number of predictors is of the non-polynomial (NP) order of the sample size. In addition, we consider a more general situation where the propensity score model may depend on predictors and needs to be estimated from data, as common in observational studies. A two-step estimation procedure is developed. In the first step, penalized regression models are fitted for the propensity score and the conditional mean of the response given predictors. In the second step, the optimal treatment regime is estimated using the penalized least squared regression with the estimated propensity score and conditional mean models obtained in the first step. There are several challenges in both numerical implementation and derivation of theoretical properties, such as weak oracle and oracle properties, for the proposed two-step estimation procedure. First,  since the posited model for the conditional mean of the response given predictors may be misspecified, the associated estimation and variable selection properties under model misspecification with NP dimensionality is not standard. Second, it is unknown how the asymptotic properties of the estimators for the optimal treatment regime obtained in the second step will depend on the estimated propensity score and conditional mean models obtained in the first step under NP dimensionality. To our knowledge, these two challenges have never been studied in the literature. 
Moreover, we estimate the value function of the estimated optimal regime and study the estimator's theoretical properties.

The remainder of the paper is organized as follows. The proposed two-step variable selection procedure for estimating the optimal treatment regime is introduced in Section 2. Section 3 and 4 demonstrate the weak oracle and oracle properties of the resulting estimators, respectively. Section 5 studies the estimator for the value function of the estimated optimal treatment regime. Simulation results are presented in Section 6. An application to a dataset from the STAR*D study is illustrated in Section 7, followed by a Conclusion Section. All the technical proofs are given in the Appendix.

\section{Method}
 Let $Y$ denote the response, $A \in \mathcal{A}$ denote the treatment received, where $\mathcal{A}$ is the set of available treatment options, and $X$ denote the baseline covariates including constant one. For demonstration purpose, we focus on a binary treatment regime, i.e., $\mathcal{A}=\{0,1\}$, with $0$ for the standard treatment and $1$ for the new treatment.
We consider the following semiparametric  model:
\begin{equation}\label{eyax}
Y=h_0(X)+A(\beta_0^T X)+e,
\end{equation}
where $h_0(X)$ is the unspecified baseline function, $\beta_0$ is the $p$-dimensional regression coefficients and $e$ is an independent error with mean $0$ and variance $\sigma^2$. Under the assumptions of stable unit treatment value (SUTVA) and no unmeasured confounders \citep{rubin1974}, it can be shown that the optimal treatment regime $d^{\mbox{\tiny{opt}}}(x)$ for patients with baseline covariates $X=x$ takes the form
\begin{eqnarray*}
	I\left(\Mean(Y|X=x, A=1)-\Mean(Y|X=x,A=0)>0\right)=I{(\beta_0^T x>0)},
\end{eqnarray*}
where $I{(\cdot)}$ is the indicator function. 

Our primary interest is in estimating the regression coefficients $\beta_0$ defining the optimal treatment regime. Let $\pi(x)= P(A = 1|X = x)$ be the propensity score. We assume a logistic regression model for $\pi(x)$:
\begin{eqnarray}\label{pi}
\pi(x, \alpha_0) = \exp(x^T \alpha_0) / [1 + \exp(x^T \alpha_0)],
\end{eqnarray}
with $p$-dimensional parameter $\alpha_0$. Here, we allow the propensity score to depend on covariates, which is common in observational studies and the parameters $\alpha_0$ can be estimated from the data. For randomized clinical trials, $\pi(x, \alpha_0)$ is a constant. We assume the majority of elements in $\beta_0$ and $\alpha_0$ are zero and refer to the support supp$(\beta_0)$, supp$(\alpha_0)$ as the true underlying sparse model of the indices.

Consider a study with $n$ subjects. Assume that the design matrix $X=(x_1, \dots, x_n)^T$ is deterministic. The observed data consist of $\{(Y_i, A_i, x_i): i=1,\cdots,n\}$. Here, we consider the case that the dimensionality $p$ is of NP order of the sample size $n$.
Define $\mu(x) = h_0(x)+\pi(x, \alpha_0) x^T \beta_0$,  the conditional mean of the response given covariates $X = x$.
We propose the following two-step estimation procedure to estimate the optimal treatment regime.
In the first step, we posit a model $\Phi(x, \theta)$ for the conditional mean function $\mu(x)$, and consider the penalized estimation for the propensity score and conditional mean models as follows.

Denote $\hat{\alpha}$ to be the $\arg\min$ of the following loss function
\begin{eqnarray}\label{alpha}
\frac{1}{n}\displaystyle \sum_{i=1}^n \left[\log \{1+\exp(x_i^T \alpha)\} - A_i x_i^T \alpha\right] + \sum_{j=1}^p\rho_1(|\alpha^j|, \lambda_{1n}),
\end{eqnarray}
and
\begin{eqnarray}\label{theta}
\hat{\theta} = \arg\min_{\theta} \frac{1}{n}\displaystyle \sum_{i=1}^n\{Y_i - \Phi(x_i, \theta)\}^2 + \sum_{j=1}^q \rho_2(|\theta^j|, \lambda_{2n}),
\end{eqnarray}
where $\alpha^j$ and $\theta^j$ refers to the $j$th element in $\alpha$ and $\theta$, $q$ is the dimension of $\theta$, and $\rho_1$ and $\rho_2$ are folded concave penalty functions with the tuning parameters $\lambda_{1n}$ and $\lambda_{2n}$, respectively. Note that the posited model $\Phi(x, \theta)$ may be misspecified.

Define $\hat{\Phi}_i=\Phi(x_i, \hat{\theta})$ and $\hat{\pi}_i=\pi(x_i, \hat{\alpha})$. In the second step, we consider the following penalized least square estimation:
\begin{eqnarray}\label{beta}\qquad
\hat{\beta} = \arg\min_{\beta}\frac{1}{n} \sum_{i=1}^n \{Y_i - \hat{\Phi}_i - (A_i - \hat{\pi}_i)\beta^T x_i\}^2 + \sum_{j=1}^p\rho_3(|\beta^j|, \lambda_{3n}),
\end{eqnarray}
where $\rho_3$ is a folded-concave penalty function with the tuning parameter $\lambda_{3n}$. Here the folded-concave penalty functions  $\rho_1$, $\rho_2$ and $\rho_3$ are assumed to satisfy the following condition: 

\begin{cond}\label{condfoldedconcave}
$\rho(t, \lambda)$ is increasing and concave in $t \in \left[0, \infty \right)$, and has a continuous derivative $\rho{'}(t, \lambda)$ with $\rho{'}(0+, \lambda) > 0$. In addition, $\rho{'}(t, \lambda)$ is increasing in $\lambda \in \left[0, \infty \right)$ and $\rho{'}(0+, \lambda)$ is independent of $\lambda$.
\end{cond}
\section{Non-asymptotic weak oracle properties}
In this section we show that the proposed estimator enjoys the weak oracle property, that is, $\hat{\alpha}$, $\hat{\beta}$ and $\hat{\theta}$ defined in (\ref{alpha})-(\ref{beta}) are sign consistent with probability tending to $1$, and are consistent with respect to the $L_\infty$ norm. Weak oracle properties of $\hat{\theta}$ are established in the sense that it converges to some least false parameter $\theta^\star$ when the main effect model is misspecified.

Theorem \ref{thmweakoracle} provides the main results. Some regularity conditions are further discussed in subsections \ref{3.2} and \ref{3.3}. A major technical challenge in deriving weak oracle properties of $\hat{\beta}$ is to analyze the deviation in (\ref{deviation}), for which we develop a general empirical process result in Proposition \ref{propdeviation}. This result  is important in its own right and can be used in analyzing many other high-dimensional semiparametric models where the index parameter of an empirical process is a plug-in estimator. The following notation is introduced to simplify our presentation.

Let $X_{1\alpha}$ and $X_{2\alpha}$, $X_{1\beta}$ and $X_{2\beta}$ be the submatrices of the design matrix $X$ formed by columns in $M_\alpha=\mbox{supp}(\alpha_0)$, $M_\alpha^c$, $M_\beta=\mbox{supp}(\beta_0)$ and $M_\beta^c$, respectively, where $M_\alpha^c$, $M_\beta^c$ stand for the complements of $M_\alpha$, $M_\beta$. Assume each $x^j$, which is the $j$th covariate in $X$, is standardized such that $||x^j||_2 = \sqrt{n}$, where $\| \cdot\|_p$ stands for the $L_p$ norm of vectors or matrices. 

Let $\Phi(\theta) = [\Phi(x_1, \theta), \dots, \Phi(x_n, \theta)]^T,$ $\phi(\theta)=[\phi^1(\theta), \dots, \phi^q(\theta)]$ denote its Jacobian matrix. The derivatives are taken componentwisely, i.e., $$\phi^l(\theta)= \left(\phi^l(x_1, \theta), \dots, \phi^l(x_n, \theta)\right),$$ for all $l=1,\dots,q$. Denote $\phi_1(\theta)$ to be a sub-matrix of $\phi(\theta)$ formed by column in $M_\theta=supp(\theta^\star)$, $\phi_2(\theta)$ formed by columns in $M_\theta^c$. We denote $\Phi(\theta^\star)$ and $\phi( \theta^\star)$ as $\Phi$ and $\phi$ when there's no confusion. 
We  use a short-hand $\hat{\Phi}$, $\hat{\phi}$ for $\Phi(\hat{\theta})$, $\phi(\hat{\theta})$. Henceforth we also write $\Phi(\theta)$, $\phi(\theta)$ with $\theta=(\theta_1^T, 0^T)^T$ as $\Phi(\theta_1)$, $\phi(\theta_1)$ for convenience.

Let $\mathbf{1}$ denote a vector of ones, $\mbox{E}$ denote the identity matrix, $O$ denote the zero matrix consisting of all zeros. For any matrix $\Psi$, let $P(\Psi)$ denote the projection matrix $\Psi (\Psi^T \Psi)^{-1} \Psi^T$. For any vector $a$, $b$, let $``\circ"$ denote the Hadamard product: $a\circ b=(a_{1}b_{1}, \dots, a_{n}b_{n})^T$, $|a|=(|a_1|, \dots, |a_n|)^T$ and  $\mbox{diag}(a)$ as the diagonal matrix with elements of vector $a$. Denote $||Y||_{\psi_m}$ as the Orlicz norm of a random variable $Y$, $$\inf_u \left\{u>0:\Mean \exp\left(\frac{|Y|}{u}\right)^m\le 2\right\},$$ for any $m\ge 1$.


\subsection{Weak oracle properties}\label{3.1}
We assume that the number of covariates $p$ and $q$ satisfy $\log p = O(n^{1-2d_\beta})$ and $\log q = O(n^{1-2d_\theta})$ for some $d_\beta$ and $d_\theta \in (0, \frac{1}{2})$, respectively.
\begin{thm}[Weak oracle property]\label{thmweakoracle}
	Assume that Conditions \ref{condregudesignmatrix}, \ref{condlambda} in \eqref{A.1}, Condition \ref{condPhi} in \eqref{3.3} hold, $\max_i ||e_i||_{\psi_1} < \infty$, 
	then there exist local minimizers $\hat{\alpha}$, $\hat{\theta}$ and $\hat{\beta}$ of the loss functions \eqref{alpha}, \eqref{theta}, and \eqref{beta} respectively, such that with probability at least $1-\bar{c}/(n+p+q)$:
	\begin{enumerate}
		\item [(a)] $\hat{\alpha}_2 = 0, \hat{\beta}_2 = 0, \hat{\theta}_2 = 0$,
		\item [(b)] $||\hat{\alpha}_1-\alpha_1||_\infty=O(n^{-\gamma_\alpha}\log n), ||\hat{\beta}_1-\beta_1||_\infty=O(n^{-\gamma_\beta}\log n), ||\hat{\theta}_1-\theta_1||_\infty=O(n^{-\gamma_\theta}\log n)$,
	\end{enumerate}
	where $\gamma_\alpha,\gamma_\beta,\gamma_\theta\le 1/2$ are some positive constants, $\bar{c}$ is some positive constant, $\hat{\alpha}_1$, $\hat{\alpha}_2$, $\hat{\beta}_1$, $\hat{\beta}_2$, $\hat{\theta}_1$ and $\hat{\theta}_2$ are sub-vectors formed by components in  $M_\alpha$, $M_\beta$, $M_\theta$ and their complements. 
\end{thm}

\begin{remark} In Theorem \ref{thmweakoracle}, part $(a)$ corresponds to the sparse recovery while $(b)$ gives the estimators' convergence rates. Weak oracle property of $\hat{\alpha}$ directly follows from Theorem 2 in \cite{fan2011}. However, to prove this property of $\hat{\beta}$ requires further efforts, to account for the variability due to plugging in $\hat {\theta} $ and $\hat {\alpha}$.
 $L_\infty$ convergence rate of $\hat{\alpha}_1$ as well as the nonsparsity size $s_\alpha$, play an important role in determining how fast $\hat{\beta}_1$ converges.  
 \end{remark}

\begin{remark}
The convergence rate of $\hat{\theta}_1$ will not affect that of $\hat{\beta}$.  This is because we require the posed propensity score model to be correct, the estimation of $\beta$ is robust with respect to the model misspecification of the main effect parameters $\theta$. Simulation results also validate our theoretical findings.
\end{remark}


\subsection{The misspecified function}\label{3.2}
We discuss conditions on misspecified function $\Phi$. Assume $\Phi$ has second order continuous derivatives. To establish weak oracle property of $\hat{\theta}$, we assume the existence and the sparsity of a least false parameter $\theta^\star$, the population parameter in the working model under the true distribution. Without loss of generality, assume the support of $\theta^\star$, $\mbox{supp}(\theta^\star)=\{1,\dots,s_\theta\}$, i.e., $\theta^\star = ({\theta_1^\star}^T, {\theta_2^\star}^T)^T$ with $\theta_1^\star$ the first $s_\theta$ nonzero component while $\theta_2^\star=0$. 

\begin{cond}\label{condPhi} We assume the following conditions:
\begin{eqnarray}
\label{phimu1}&&\displaystyle \sup_{\delta \in H_\theta}||\{\phi_1(\delta)^T \phi_1(\delta)\}^{-1}\phi_1(\delta)^T (\mu - \Phi)||_\infty = O(n^{-\gamma_\theta} \log n),\\
\label{phimu2}&&\displaystyle \sup_{\delta \in H_\theta}||\phi_2(\delta)^T [E-P\{{\phi_1}(\delta)\}] (\mu - \Phi)||_\infty=O(n^{1-d_\theta}\sqrt{\log n}),\\
\label{phiphiinfty}&&\sup_{\delta \in H_\theta}||\{\phi_1(\delta)^T \phi_1(\delta)\}^{-1}||_\infty = O(\frac{b_\theta}{n}),\\
\label{phi2phi1infty}&&\sup_{\delta \in H_\theta}||\phi_2(\delta)^T \phi_1(\delta) \{\phi_1(\delta)^T \phi_1(\delta)\}^{-1}||_\infty \le \min\left\{C \frac{\rho^{'}_3(0+)}{\rho^{'}_3(d_{n\theta})}, O(n^{a_3})\right\},\\
\label{phi2norm}&&\displaystyle \max_{l=1}^q ||\phi^l \circ (\mathbf{1}+ |X\beta_0|)||_2=O(\sqrt{n}),\\
\label{dphi2norm}&&\displaystyle \max_{l=1}^q \sum_{k=1}^{s_\theta} \sup_{\delta \in H_\theta} ||\frac{\partial\phi^l(\delta)}{\partial \theta_k}\circ (\mathbf{1}+ |X\beta_0|)||_2=O(\frac{n^{\frac{1}{2}+\gamma_\theta}}{\sqrt{s_\theta} \log n}),\\
\label{lamphidphi}&&\sup_{\delta_1 \in H_\theta}\sup_{\delta_2 \in H_\theta}\displaystyle \max_{l=1}^q \lambda_{\max} \left(\frac{\partial (|\phi^l(\delta_1)|)^T \phi_1(\delta_2)}{\partial \theta_1}\right) = O(n),
\end{eqnarray}
where $0\ge a_3\le 1/2$ some positive constant, $d_{n\theta}=\min_{j\in M_\theta} |{\theta^\star}^j|$ the minimum half signal,
$H_\theta = \left\{\delta\in \mathbb{R}^{s_\theta}: ||\delta - \theta^\star||_\infty=O(n^{-\gamma_\theta}\log n)\right\}$. If the response is unbounded, we require
\begin{eqnarray}
\label{phimax}&&\displaystyle \max_{l=1}^q (||\phi^l||_\infty) = o(n^{d_\theta}/\sqrt{\log n}),\qquad\qquad\qquad\qquad\qquad\qquad\qquad\quad~~
\end{eqnarray}
and the right-hand side of \eqref{dphi2norm} shall be modified to $O(n^{\frac{1}{2}+\gamma_\theta}/\sqrt{s_\theta} \log^2 n)$.
\end{cond}

\begin{remark}
Conditions \eqref{phimu1} and \eqref{phimu2} are key assumptions determining the degree of model misspecification. 
Condition \eqref{phimu1} requires that the posited working model $\Phi$ can provide a good approximation for $\mu$. In that case, the residual $\mu-\Phi$ will be orthogonal to the jacobian matrix $\phi_1$ and the left-hand side of \eqref{phimu1} will be small. There's a trade-off between the simplicity of $\Phi$ and the accuracy in terms of approximating $\mu$. 
Condition \eqref{phimu2}, which measures $\Phi$'s complexity, automatically holds for the oracle submodel of $\Phi$. 
\end{remark}

\begin{remark}
Conditions \eqref{dphi2norm} and \eqref{lamphidphi} put constraints on the derivatives of $\phi$, requiring the misspecified function to be smooth. The right-hand side order in \eqref{dphi2norm} is not too restrictive since we require $n^{\gamma_\theta} \gg s_\theta \log n$. 
\end{remark}

Three common examples of the main-effect function $\Phi$ are provided below to examine the validity of Condition \ref{condPhi}.


\begin{exam}
	Set $\Phi=0$. Then, no model is needed for $\Phi$. It is easy to check that Condition 3.1. is satisfied.
\end{exam}

\begin{exam}
	 When a linear model is specified, i.e., $\Phi(x,\theta)=x^T \theta$, conditions \eqref{dphi2norm} and \eqref{lamphidphi} are automatically satisfied since the second-order derivative of $\Phi$ vanishes. If we set $d_\theta=d_\beta$, \eqref{phimax} holds.  Condition \eqref{phimu1} also satisfies automatically. 
	 Condition \eqref{phimu2} becomes
	 \begin{eqnarray}
	 \label{x2thetamu}||X_{2\theta}^T \{I-P(X_{1\theta})\} \mu||_\infty&=&O(n^{1-d_\theta}\sqrt{\log n}),
	 \end{eqnarray}
	 each element in the left-hand side vector in \eqref{x2thetamu} can be viewed as the inner product of the residuals obtained by fitting $X_{1\theta}$ on each noise variable in $X_{2\theta}$ and those fitted by regressing $X_{1\theta}$ on $\mu$. When $\mu$ depends only on $X_{1\theta}$, \eqref{x2thetamu} holds with $d_\theta=d_\beta$ for the Gaussian linear model.
\end{exam}


\subsection{The design matrix}\label{3.3}
We provide the technical conditions on the design matrix as follows.
\begin{cond}\label{condregu}
	Assume that
	\begin{eqnarray}\qquad
	\label{xxa1norm}&&\sup_{\delta \in H_\theta}||B_{n\beta}^{-1} X_{1\beta}^T W(\delta) \Delta X_{1\alpha} B_{n\alpha}^{-1}||_\infty = O(\frac{b_{\alpha\beta}}{n}), \\
	\label{x2bx1}&&\sup_{\delta \in H_\theta}||X_{2\beta}^T W_\beta W(\delta) X_{1\alpha} B_{n\alpha}^{-1}||_\infty\ = \min\left\{o\left(\frac{\lambda_{2n}\rho^{'}_2(0+)}{\lambda_{1n}\rho^{'}_1(d_{n\beta})}\right), O(n^{a_2})\right\},\\
	\label{xmu2norm}&&\max_{j=1}^p ||W(\theta_1^\star) x^j||_2 = O(\sqrt{n}),\\
	\label{xxaxb2norm}&&\max_{j=1}^p \sum_{k\in M_\alpha}||x^k \circ x^j \circ (X\beta_0) ||_2 = O(\frac{n^{1/2+\gamma_\alpha}}{\log n}),\\
	\label{xx2norm}&&\max_{j=1}^p \sum_{k\in M_\beta}||x^j \circ x^k||_2 = O(\frac{n^{1/2+\gamma_\beta}}{\log n}),\\
	\label{xphi2norm}&&\max_{j=1}^p \sum_{l\in M_\theta} \sup_{\delta \in H_\theta}||x^j \circ \phi^l(\delta)||_2 = O(\frac{n^{1/2+\gamma_\theta}}{\sqrt{s_\theta}\log^3 n}),\\
	\label{lamxmu-phi}&&\sup_{\delta \in H_\theta}\max_{j=1}^p \lambda_{\max}[X_{1\alpha}^T \diag(|W(\delta) x^j|)X_{1\alpha}] = O(n),\qquad\qquad\qquad\\
	\label{lamxxbeta}&&\max_{j=1}^p \lambda_{\max}[X_{1\alpha}^T \diag |x^j \circ (X\beta_0)|X_{1\alpha}] = O(n),
	\end{eqnarray}
	where
	\begin{eqnarray*}
		&&W(\delta)=\diag[\mu-\Phi(\delta)],\quad B_{n\alpha}=X_{1\alpha}^T \Delta X_{1\alpha}, \quad B_{n\beta}=X_{1\beta}^T \Delta X_{1\beta},\\
		&&W_\beta=\Delta-\Delta^{\frac{1}{2}} P(\Delta^{\frac{1}{2}} X_{1\beta})\Delta^{\frac{1}{2}}.
	\end{eqnarray*}
\end{cond}

The sequence $b_{\alpha \beta}$ in (\ref{xxa1norm}) shall satisfy
\begin{eqnarray*}
b_{\alpha \beta} = \min \left\{o(n^{\frac{1}{2}-\gamma_\beta} \sqrt{\log n}), o(n^{2\gamma_\alpha-\gamma_\beta}/s_\alpha \log n)\right\}.
\end{eqnarray*}

\begin{remark}
	Conditions \eqref{xxa1norm} and \eqref{x2bx1} control the impact of the deviation of the estimated propensity score from its true value on $\hat{\beta}$, thus are not needed when the propensity scores are pre-specified. By the definition of $W(\delta)$, magnitudes of the left-hand side in these two conditions depend on how accurate the misspecified function models $\mu$ and how fast $\hat{\theta}$ converges to the least square estimator. The sequence $b_{\alpha\beta}$ in \eqref{xxa1norm} can converge to $0$ when $X_{1\beta}$ and $X_{1\alpha}$ are weakly correlated. Even in the most extreme case where $X_{1\alpha}=X_{1\beta}$, i.e, the propensity score model and outcome regression model share the same important variables, $b_{\alpha\beta}$ usually would not exceed the order $O(b_\beta)$. Each element in the left-hand side of \eqref{x2bx1} is the multiple regression coefficient of the corresponding variable in  $X_{2\beta}$ on $W(\delta) X_{1\alpha}$, using weighted least squares with weights $\pi \circ (1-\pi)$, after adjusted by $X_{1\beta}$, which characterize their weak dependence given $X_{1\beta}$. These two conditions are generally weaker than those imposed by \cite{fan2011} (Condition 2), and are therefore more likely to hold.
\end{remark}

\begin{remark}
The right-hand side in \eqref{xphi2norm} can be relaxed to $O(n^{1/2+\gamma_\theta}/\log n)$ when using the linear model. The additional term $\sqrt{s_\theta}$ is due to the penalty on the complexity of the main effect model.  
This condition typically controls the deviation
\begin{eqnarray}\label{deviation}
||Z^T \{\Phi-\Phi(\hat{\theta})\}||_\infty=O_p(\sqrt{\log p\log n}),
\end{eqnarray}
where $Z=\diag(A-\pi)X$. 
A common approach to bound the deviation is to utilize the classical Bernstein's inequality. However this approach does not work here, because the indexing parameter in the process $\Phi(\cdot)$ in (\ref{deviation}) is an estimator.
To handle this challenge, we bound the left-hand side in (\ref{deviation}) by
\begin{eqnarray*}
\sup_{\theta_1,\theta_2\in H_\theta}||Z^T \{\Phi(\theta_1)-\Phi(\theta_2)\}||_\infty.
\end{eqnarray*}
A general theory that covers the above result is provided in Proposition \ref{propdeviation}.
\end{remark}

\begin{remark}
Conditions \eqref{lamxmu-phi} and \eqref{lamxxbeta} aim to control the $L_\infty$ norm of the quadratic term of the Taylor series as a function of $\hat{\alpha}$, expanded at $\alpha_0$. Similar to \eqref{xxa1norm} and \eqref{x2bx1}, the two conditions are not needed when $\alpha_0$ is known to us.
\end{remark}

\subsection{Deviation}\label{3.5}
%
Let $a_{ij}(h), i=1,\dots,n,j=1,\dots,J$ be arbitrary deterministic functions, continuously differentiable with respect to an $S$-dimensional parameter $h$. We restrict the domain to the hypercube $H$ centered at some fixed vector $h_0$, $H=\{h:||h-h_0||_\infty=\delta\}$. For $s=1,\ldots, S$, define $b_{ij}^s(h)=\frac{\partial a_{ij}(h)}{\partial h_s}$ where $h_s$ is the $s$th component of $h$, $b_j^s(h)=[b_{1j}^s(h), \dots, b_{nj}^s(h)]^T$, $A(h)$ the $J \times n$ matrix, $\{a_{ij}(h)\}$.

\begin{prop}\label{propdeviation} Let $z=(z_1,\dots,z_n)^T$ be the $n$-dimensional independent random vector with mean $0$, $\max_i ||z_i||_{\psi_1}\le \omega$, $\max_i z_i^2=v_0^2$, let
	\begin{eqnarray*}
	d_n=\max_{j=1,\dots,J} \sum_{s=1}^S \sup_{h\in H} ||b^s_j(h)||_2,
	\end{eqnarray*}
	then
	\begin{eqnarray}\label{prop1eq1}
	\qquad\Mean \left(\sup_{h_1,h_2 \in H}||\{A(h_1)-A(h_2)\}^T z||_\infty \right)\le c_0 \delta \sqrt{S\log (J)}d_n\omega\log n.
	\end{eqnarray}
	Further, for any $t\ge 3\delta v_0 d_n$, we have
	\begin{eqnarray}\nonumber
	\prob\left(\sup_{h_1,h_2 \in H}||\{A(h_1)-A(h_2)\}^T z||_\infty>t\right)\\\label{prop1eq2}
	\le \frac{8}{n}+4J\exp\left(-\frac{c_0t^2}{\delta^2 S d_n^2 \omega^2\log^4 n}\right),
	\end{eqnarray}
	for some constant $c_0$. When $z_i$'s are bounded, the term $\omega\log n$ in \eqref{prop1eq1} and \eqref{prop1eq2} can be removed.
\end{prop}

\begin{remark} Proposition \ref{propdeviation} shows that the magnitude of $$\sup_{h_1,h_2 \in H}||\{A(h_1)-A(h_2)\}^T z||_\infty$$ is determined by the $L_2$ constraints on the derivatives, the number of sums $J$, the dimension of the parameter $S$ and the diameter of the region $\delta$.
\end{remark}

\section{Oracle properties}
In this section we study the oracle property of the estimator $\hat \beta$. 
 We assume that  $\max(s_\alpha, s_\beta) \ll \sqrt{n}$ and $n^{\gamma_\theta} \gg s_\theta \log n$. 
 The convergence rates of the estimators are established in Section \ref{4.1} and their asymptotic distributions are provided in Section \ref{4.2}. Technical conditions  are discussed in Section \ref{4.3}.
\subsection{Rates of convergence}\label{4.1}
\begin{thm}\label{thmrateofconvergence}
	Assume that Conditions \ref{condfoldedconcave}, \ref{condPhi}, \ref{condregudesignmatrix2}, \ref{condlambda2} and \ref{condrateofconvergence} hold and $\max_i ||e_i||_{\psi_1} < \infty$. Constraints on $b_\theta$, $d_\theta$,$d_{n\theta}$ and $\lambda_{3n}$ are same as in Theorem \ref{thmweakoracle}. Further assume $\max(l_1, l_2) < \frac{1}{2}$ with $s_\alpha = O(n^{l_1})$, $s_\beta = O(n^{l_2})$, and $n^{\gamma_\theta} \gg s_\theta \log n$. Then there exists a strict local minimizer $\hat{\beta}$ of the loss function \eqref{beta}, $\hat{\alpha}$ of \eqref{alpha}, such that $\hat{\alpha}_2 = 0$,  $\hat{\beta}_2 = 0$ with probability tending to $1$ as $n \to \infty$, and $||\hat{\alpha}_1 - \alpha_1||_2 = O(\sqrt{s_\alpha+s_\beta}n^{-1/2})$,  $||\hat{\beta}_1 - \beta_1||_2 = O(\sqrt{s_\beta}n^{-1/2})$. 
\end{thm}

\begin{remark} 
	We note that when establishing the oracle property of $\hat{\beta}$, only the weak oracle property of $\hat{\theta}$ is required. This is due to the robustness of the A-learning methods and the fact that the propensity score is correctly specified.
\end{remark}

\begin{remark}
	Precision of $\hat{\beta}_1$ is affected by that of $\hat{\alpha}_1$, since $||\hat{\beta}_1-\beta_1||_2$ is at least the same order of magnitude as $||\hat{\alpha}_1-\alpha_1||_2$. When the propensity score is known, convergence rate of $\hat{\beta}_1$ is improved to $\sqrt{s_\beta/n}$.
\end{remark}



\subsection{Asymptotic distributions}\label{4.2}
To establish the weak convergence of the estimators, we define $\Sigma_{12}$ and $\Sigma_{22}$ as
	\begin{eqnarray*}
		\Sigma_{12} &=& 2B_{n\alpha}^{-1/2} X_{1\alpha}^T \Delta W X_{1\beta} B_{n\beta}^{-1/2},\\
		\Sigma_{22} &=& B_{n\beta}^{-1/2} X_{1\beta}^T W \Delta^{1/2}(E- P_{\Delta^{\frac{1}{2}}X_{1\alpha}})\Delta^{1/2} W X_{1\beta} B_{n\beta}^{-1/2},
	\end{eqnarray*}
	where $W$ is a shorthand for $W(\theta_1^\star)$.

\begin{thm}[Oracle property]\label{thmoraclebeta}
	Under conditions in theorem \ref{thmrateofconvergence} and Condition \ref{condoralceproperty}, if $\max(s_\alpha, s_\beta)=o(n^{1/3})$, the right-hand side of \eqref{xphi2norm} is strengthened to $O(n^{\frac{1}{2}+\gamma_\theta}/\sqrt{s_\beta s_\theta\log^3 n})$, 
	as $n\to \infty$, with probability tending to $1$, $\hat{\alpha} = (\hat{\alpha}_1^T, \hat{\alpha}_2^T)^T$, $\hat{\beta} = (\hat{\beta}_1^T, \hat{\beta}_2^T)^T$ in Theorem 2 satisfy
	\begin{enumerate}
		\item [(a)] $\hat{\alpha}_2 = 0$, $\hat{\beta}_2 = 0$,
		\item [(b)] $[A_{1n} B_{n\alpha}^{1/2} (\hat{\alpha}_1 - \alpha_1), A_{2n} B_{n\beta}^{1/2} (\hat{\beta}_1 - \beta_1)]$ is asymptotically normally distributed with mean $0$, covariance matrix $\Omega$, which is the limit of
		\begin{eqnarray*}
			\left(\begin{array}{cc}
				A_{1n} A_{1n}^T & A_{1n} \Sigma_{12} A_{2n}^T \\
				A_{2n} \Sigma_{21} A_{1n}^T & \sigma^2 A_{2n} A_{2n}^T + A_{2n} \Sigma_{22} A_{2n}^T\\
			\end{array}\right),
		\end{eqnarray*}
	\end{enumerate}
	where $A_{1n}$ is a $q_1 \times s_\alpha$ matrix and $A_{2n}$ is a $q_2 \times s_\beta$ matrix such that $$\lambda_{\max}(A_{1n} A_{1n}^T)=O(1), \quad \lambda_{\max}(A_{2n} A_{2n}^T)=O(1).$$
	
\end{thm}
We note that conditions on the smoothness of the misspecified function \eqref{xphi2norm} is strengthened. 
To better understand the above theorem, we provide the following two corollaries. The first corollary gives the limiting distribution when we specify both the propensity score and main-effect model while the second one corresponds to case when the propensity score is known in advance.

\begin{coro}\label{corobetabaseline}
	Under conditions of Theorem \ref{thmoraclebeta}, when we correctly specify the main-effect model, i.e., $\mu=\Phi$, $A_{1n} \hat{\alpha}_1$ and $A_{2n} \hat{\beta}_1$ are jointly asymptotically normally distributed, with the covariance matrix $\Omega^\prime$, which is the limit of the following matrix,
	\begin{eqnarray*}
		\left(\begin{array}{cc}
			A_{1n}^T A_{1n} &  O\\
			O & \sigma^2 A_{2n}^T A_{2n}\\
		\end{array}\right).
	\end{eqnarray*}
\end{coro}
\begin{remark}
Comparing the results in Corollary \ref{corobetabaseline} and in Theorem \ref{thmoraclebeta}, the term $A_{2n}^T \Sigma_{22} A_{2n}$ accounts for the partially specification of model \eqref{eyax}. In the most extreme case where we correctly specify $\Phi$, $\hat{\beta}$ will achieve its minimum variance and is independent of $\hat{\alpha}_1$. In general, we can gain efficiency by posing a good working model for $\Phi$. Numerical studies also suggest that a linear model such as $\Phi=X\theta$ is preferred compared to the constant model. This is in line to our theoretical justification since $W$ is a diagonal matrix with the $i$th diagonal element $\mu_i-\Phi_i$.
\end{remark}
\begin{coro}\label{corobetaalphaknown}
	When the propensity score is known, under conditions of Theorem 3 with all $\hat{\alpha}$'s  replaced by $\alpha_0$, then with probability tending to $1$ as $n \to \infty$, $A_{2n} B_{n\beta}^{1/2}(\hat{\beta}_1-\beta_1)$ is asymptotically normally distributed with mean $0$, covariance matrix $\Omega{''}$ which is the limit of
	\begin{eqnarray*}
	\sigma^2 A_{2n}^T A_{2n}+ A_{2n}^T \Sigma_{22}^\prime A_{2n},
	\end{eqnarray*}
	where
	$$\Sigma_{22}^\prime=B_{n2}^{-1/2}X_{1\beta}^T W \Delta W X_{1\beta} B_{n2}^{-1/2}.$$ 
\end{coro}

\begin{remark}
	An interesting fact implied by Corollary \ref{corobetaalphaknown} is that the asymptotic variance of $\hat{\beta}_1$ will be smaller than that of the same estimator had we known the propensity score in advance. A similar result is given in the asymptotic distribution of the mean response for the value function in the next section. This is in line with the semiparametric theory in fixed $p$ case where the variance of AIPWE would be smaller when we estimate the parameter in the coarsening probability model, even if we know what the true value is \citep[see Chapter 9 in][]{butch2006}. By doing so, we can actually borrow information from the linear association between covariates in $WX_{1\beta}$ and those in $X_{1\alpha}$. 
\end{remark}

\subsection{Technical conditions}\label{4.3}
\begin{cond}\label{condrateofconvergence} In addition to \eqref{lamxmu-phi} and \eqref{lamxxbeta} in Condition \ref{condregu}, assume that the right-hand side of \eqref{xphi2norm} is strengthened to $O(n^{\frac{1}{2}+\gamma_\theta}/\sqrt{s_\theta\log^3 n})$, and the following conditions hold,
\begin{eqnarray}
\label{xx1a2norm}&&\displaystyle \sup_{\delta \in H_\theta}||B_{n\beta}^{-1/2} X_{1\beta}^T W(\delta) \Delta X_{1\alpha} B_{n\alpha}^{-1/2}||_2 = O(\frac{\sqrt{s_\beta}}{\sqrt{s_\alpha}}),\\
\label{xx1a2norm2}&&\displaystyle \sup_{\delta \in H_\theta}||X_{2\beta}^T W_\beta W(\delta) X_{1\alpha}||_{2,\infty} = O(n\frac{\sqrt{s_\beta}}{\sqrt{s_\alpha}}),\\
\label{xx2norm2}&&\max_{j=1}^p \max_{k\in M_\beta} ||x^j \circ x^k||_2 = O(\sqrt{n}),\\
\label{xmux1a2norm2}&&\max_{j=1}^p \max_{k\in M_\alpha}||x^j\circ x^k \circ (X \beta_0)||_2 = O(\sqrt{n}),\\
\label{tracexmu}&& tr\left[X_{1\beta}^T W(\theta_1^\star) \Delta W(\theta_1^\star) X_{1\beta}\right] = O(s_\beta n).
\end{eqnarray}
\end{cond}
\begin{remark}
Similar to the interpretation of \eqref{xxa1norm} and \eqref{x2bx1}, \eqref{xx1a2norm} corresponds to a notion of weak dependence between variables in $X_{1\alpha}$ and $X_{1\beta}$ while \eqref{xx1a2norm2} require $X_{2\beta}$ and $X_{1\alpha}$ are weakly correlated after adjusted by $X_{1\beta}$. Validity of these two conditions are related with the convergence rate of $\hat {\theta}_1$ and the magnitude of model misspecification $||\mu-\Phi||_\infty$. Besides, it can be verified that \eqref{xx2norm2}-\eqref{tracexmu} hold with large probability when the baseline covariates possesses subgaussian tail.
\end{remark}

\begin{cond}\label{condoralceproperty}
Assume that
\begin{eqnarray}
\label{penalty3}&&\displaystyle \lambda_{1n} \bar{\rho}_1(d_{n\alpha}) = o(s_\alpha^{-1/2} n^{-1/2}),\quad \lambda_{2n} \bar{\rho}_2(d_{n\beta}) = o(s_\beta^{-1/2} n^{-1/2}),\\
\label{lyapunov3}&&\displaystyle \sum_{i=1}^n (x_{1\alpha i}^T B_{n\alpha}^{-1} x_{1\alpha i})^{3/2} \to 0,\quad \sum_{i=1}^n (x_{1\beta i}^T B_{n\beta}^{-1} x_{1\beta i})^{3/2} \to 0,\\
\label{lyapunov3mu-phi}&&\displaystyle \sum_{i=1}^n (x_{1\beta i}^T B_{n\beta}^{-1} x_{1\beta i})^{3/2}|\mu_i-\Phi_i|^3 \to 0,\\
\label{cov3}&&\displaystyle \displaystyle \lambda_{\max} \left(B_{n\beta}^{-1/2} X_{1\beta}^T W^2 X_{1\beta} B_{n\beta}^{-1/2}\right)=O(1),\\
\label{covphi3}&&\displaystyle \sup_{\delta \in H_\theta} ||B_{n\beta}^{-1/2}X_{1\beta}^T \diag[\Phi-\Phi(\delta)] \Delta X_{1\alpha} B_{n\alpha}^{-1/2}||_2=o(1).
\end{eqnarray}
where $x_{1\alpha i}$ and $x_{1\beta i}$ stand for the $i$th row of the matrix $X_{1\alpha}$ and $X_{1\beta}$ respectively.
\end{cond}

\begin{remark}
Conditions \eqref{lyapunov3} and \eqref{lyapunov3mu-phi} are the Lyapunov conditions which guarantee the normality of $\hat{\alpha}_1$ and $\hat{\beta}_1$. Condition \eqref{cov3} puts constraints on the maximum eigenvalue of the variance-covariance matrix of $Z_1^T (\mu-\Phi)$ by requiring it to be finite. Condition \eqref{covphi3} holds when $\Phi(\delta)$ converges to $\Phi$ uniformly in terms of $L_\infty$ norm with $\delta$ in the region $H_\theta$. When $||\mu-\Phi||_\infty$ is bounded, \eqref{lyapunov3mu-phi} and \eqref{cov3} are simultaneously satisfied. 
\end{remark}

\section{Evaluation of value function}
In this section, we derive a non-parametric estimate for the mean response under the optimal treatment regime. By (\ref{eyax}), define our average population-level response under a specific regime as
\begin{eqnarray*}
V_n(\beta)&=&\frac{1}{n}\sum_{i=1}^n\Mean[Y_i|A_i=I(x_i^T \beta>0), X_i=x_i]\\
&=&\frac{1}{n}\sum_{i=1}^n[h_0(x_i)+x_i^T \beta_0 I(x_i^T \beta>0)],
\end{eqnarray*}
where the treatment decision for the $i$th patient is given as $I(x_i^T \beta>0)$. The mean response under the true optimal regime is denoted as $V_n(\beta_0)$ and it is easy to verify that $\beta_0$ is the maximizer of the function $V_n$.


Similarly as in \cite{Murphy2003}, we propose to estimate the value function $V_n(\beta_0)$ using
\begin{eqnarray}\label{vhat}
\hat{V}_n=\frac{1}{n}\sum_{i=1}^n [Y_i+x_i^T \hat{\beta} \{I(x_i^T \hat{\beta}>0)-A_i\}].
\end{eqnarray}
This estimator is not doubly robust but offers protection against misspecification of the baseline function and improved efficiency. It's not doubly robust because we require the propensity score model to be correctly specified to ensure the oracle property of $\hat{\beta}$. A key condition which guarantees asymptotic normality of (\ref{vhat}) is given as follows.

\begin{cond} Assume there exists some constant $C^\prime$, such that for all $\varepsilon>0$,
\begin{eqnarray*}
\frac{1}{n} \sum_i I(|x_i^T \beta_0|<\varepsilon)\le C^\prime \varepsilon.
\end{eqnarray*}
\end{cond}

\begin{remark}
The above condition has similar interpretation as Condition (3.3) in \cite{qian2011}, where random design were utilized. 
 Condition 7 requires that the absolute value of the average contrast function can not be too small, which together with the condition  $s_\beta=o(n^{1/4})$ ensures the following stochastic approximation condition:
\begin{eqnarray}\label{stoappx}
\sqrt{n} \sum_i x_i^T \hat{\beta} \{I(x_i^T \hat{\beta}>0)-I(x_i^T \beta_0>0)\}=o_p(1).
\end{eqnarray}
\end{remark}


\begin{thm}\label{thmvhat}
Assume that conditions in Theorem 3 hold. If Condition 7 holds and the nonsparsity size $s_\beta$ satisfies $s_\beta=o(n^{1/4})$, then with probability going to $1$, $\sqrt{n}\{\hat{V}_n-V_n(\beta_0)\}$ is asymptotically normally distributed with variance $\upsilon_0^2$, which is limit of
\begin{eqnarray}\label{variance}
\sigma^2+\sigma^2 v_n^T X_{1\beta} B_{n\beta}^{-1} X_{1\beta}^T v_n+ v_n^T X_{1\beta} B_{n\beta}^{-1/2} \Sigma_{22} B_{n\beta}^{-1/2} X_{1\beta}^T v_n,
\end{eqnarray}
where $v_n$ stands for the vector $[I(x_1^T \beta_0>0)-\pi_1, \dots, I(x_n^T \beta_0>0)-\pi_n]^T/\sqrt{n}$, and $\Sigma_{22}$ is defined in Theorem \ref{thmoraclebeta}.
\end{thm}

\begin{remark}
Note that we only need $s_\beta=o(n^{1/2})$ to guarantee the weak oracle property of $\hat{\beta}$ or $O(\sqrt{s_\beta}/\sqrt{n})$ convergence rate of $||\hat{\beta}_1-\beta_1||_2$. This condition is strengthened to $s_\beta=o(n^{1/3})$ to show the asymptotic normality of $\hat{\beta}_1$. Theorem \ref{thmvhat} further requires $s_\beta=o(n^{1/4})$ as to ensure the approximation condition \eqref{stoappx}.
\end{remark}

\begin{remark} When \eqref{stoappx} is satisfied, the asymptotic normality of $\hat{V}_n$ follows immediately from the oracle property of the estimator $\hat{\beta}_1$. The first term $\sigma^2$ in \eqref{variance} is due to variation of the error term $e_i$ while the last two terms correspond to the asymptotic variance of $\hat{\beta}_1$. 
\end{remark}
We provide a corollary here which corresponds to the case where the main-effect model is correctly specified.

\begin{coro}
In addition to the conditions in Theorem 4, if the main-effect model is correct, $\sqrt{n}\{\hat{V}_n-V_n(\beta_0)\}$ is asymptotically normally distributed with variance $\upsilon_1^2$, which is defined as the limit of
\begin{eqnarray*}
\sigma^2+\sigma^2 v_n^T X_{1\beta} B_{n\beta}^{-1} X_{1\beta}^T v_n,
\end{eqnarray*}
where $v_n$ is defined in Theorem \ref{thmvhat}.
\end{coro}

Similar to the asymptotic distribution of $\hat{\beta}_1$, the following corollary suggests that the proposed estimator is more efficient in the case when we estimate the propensity score by fitting a penalized logistic regression.

\begin{coro}
Assume the propensity score is known, and conditions in Theorem \ref{thmvhat} hold with all $\hat{\alpha}$'s replaced by $\alpha_0$, then with probability going to $1$, $\sqrt{n}\{\hat{V}_n-V_n(\beta_0)\}$ is asymptotically normally distributed with variance $\upsilon_2^2$, which is the limit of
\begin{eqnarray*}
&\sigma^2+\sigma^2 v_n^T X_{1\beta} B_{n\beta}^{-1} X_{1\beta}^T v_n+ v_n^T X_{1\beta} B_{n\beta}^{-1/2} \Sigma_{22}^\prime B_{n\beta}^{-1/2} X_{1\beta}^T v_n,
\end{eqnarray*}
with $v_n$  defined in Theorem \ref{thmvhat}, and $\Sigma_{22}^\prime$ defined in Corollary \ref{corobetaalphaknown}.
\end{coro}

By the definition of $v_n$ and the condition that $\lambda_{\max}(X_{1\beta}^T X_{1\beta})=O(n)$, the asymptotic variance will reach its minimum when 
$I(x_i^T \beta_0>0)$ is close to the propensity score. We characterize this result in the following Corollary.

\begin{coro}
	Under the conditions in Theorem \ref{thmvhat}, if we further assume that
	\begin{eqnarray*}
	\frac{1}{n} \sum_{i=1}^n \{I(X_i^T\beta_0>0)-\pi_i\}^2=o(1),
	\end{eqnarray*}
	then with probability going to $1$, $\sqrt{n}\{\hat{V}_n-V_n(\beta_0)\}$ is asymptotically normally distributed with the variance $\sigma^2$.
\end{coro}

\begin{remark}
Such a result is expected with the following intuition: 
in an observational study, if the clinician or the decision maker has a high chance to assign the optimal treatment to an individual patient, i.e., the propensity score is close to $I(x_i^T \beta_0>0)$, the variation in estimating the value function will be decreased. In other words, the more skillful the clinician or the decision maker is, the closer the observed individual response $Y_i$ approaches the potential outcome under the optimal treatment regime.
\end{remark}

\section{Numerical studies}
In this section, we evaluate the numerical performance of the proposed estimators in various settings.

\subsection{Performance of the new method}
We generate the propensity score from the logistic regression model (\ref{alpha}), with five nonzero coefficients $\alpha_0=(1.5, -1.0, 1.4, 0.8, -1.2)^T$. Following \cite{lv2011}, we choose three forms for the baseline function $h_0(x)$, including a simple linear form, a quadratic form and a complex non-linear form,

\begin{itemize}
	\item Model I: $Y=\mathbf{1}+\gamma_1^T X+A (\beta_0^T X) + \epsilon$,
	\item Model II: $Y=\mathbf{1}+0.5(\gamma_1^T X)(\gamma_2^T X)+A (\beta_0^T X) + \epsilon$,
	\item Model III: $Y=\mathbf{1}+0.5\sin(\pi\gamma_1^T X)+0.25(\mathbf{1}+\gamma_2^T X) (\mathbf{1}+\gamma_2^T X)+A  (\beta_0^T X) + \epsilon$.
\end{itemize}
The important variables we choose in the baseline function are different from those in the logistic model (\ref{alpha}) and contrast function (\ref{beta}). We allow $\alpha_0$ and $\beta_0$ to share two nonzero entries. We set $p=1000$, the nonzero components of $\gamma_1=(0.5, -0.5, 0.5, -0.5, 0.5)^T$ and $\gamma_2=(-0.5, 0.5, -0.5, 0.5, -0.5)^T$.

The number of simulations was $500$. For each data set, we generated the rows of $X$ to be $p$ independent standard normal random variables, or jointly generated as i.i.d. samples from normal random variables  with mean zero and variance  $\Sigma_0=(0.3^{|i-j|})_{i,j=1,\dots,p}$. When the covariates are independently generated, the five nonzero entries for $\alpha_0$ and $\beta_0$ are $1,2,3,4,5$ and $1,2,6,7,8$, respectively. When they follow an AR$(1)$ structure, we set the nonzero entries as $1,2,9,10,50$ for $\alpha_0$ and $1,2,15,16,100$ for $\beta_0$. Besides, we consider two choices for the magnitude of the half minimal signal of $\beta_0$. In the moderate case, we set $\beta_1$ to be $(0.8, -0.5, -0.6, 1.0, -0.6)^T$ and $(2.0, -1.3, 1.5, -1.2, 1.0)^T$ in the large magnitude setting. We also consider two different sample sizes, $n=200$ and $n=400$, which yields a total of $8$ scenarios. In each scenario, we fit a linear model for (\ref{theta}) and use SCAD penalty function with the regularization parameter chosen according to  $10$-fold cross-validation.

To evaluate the performance of the estimator, we report the $L_2$ loss of $\hat{\beta}$ and that of the first-stage estimator $\hat{\alpha}$ as well. The number of missed true variables in $X_{1\alpha}$ and $X_{1\beta}$ (denoted as FN), the number of selected variables (denoted as $\#S$) and the average percentage of making correct decisions (denoted as PCD), which is defined as $1-\sum_{i=1}^n|I(\hat{\beta}^T x_i>0)-I(\beta_0^Tx_i>0)|/n$ are also reported. In addition, we estimate $E(Y^\star(\hat{d}))$ and $E(Y^\star(d^{\tiny{opt}}))$,  the value functions of our estimated optimal treatment regime and the true optimal regime, respectively, using Monte Carlo simulations.
For example, we compute $E(Y^\star(\hat{d}))$ by generating data for $10000$ subjects from the model,
\begin{eqnarray*}
Y=h_0(x)+Ax^T\beta_0+\epsilon,
\end{eqnarray*}
where $A$ is determined by the estimated optimal treatment regime $I(\hat{\beta}^T x>0)$. Taking the average of the $10000$ values, we report the averages of mean responses over $500$ replications as well as their standard deviations.

\begin{table}[htbp]
\caption{$L_2$, $\#S$, FN of $\hat{\beta}$ and $\hat{\alpha}$ (in parentheses), PCD, average mean and standard deviation (in parentheses) of $Y^\star(\hat{d})$ and $Y^\star(d^{\tiny{opt}})$ over 500 simulations for all three models where $n=200$, $400$}
\begin{tabular}{cccccc}\hline
& Measures & n & Model I & Model II & Model III\\ \hline
moderate & $L_2$ loss & 200 &0.868(1.310) &1.361(1.308) &1.294(1.240)\\
magnitudes&    &400 &0.504(0.726) &0.987(0.723) &0.779(0.717)\\
for $d_{n\beta}$& FN &200&0.572(0.114) &2.318(0.096) &1.680(0.134)\\
with&    &400  &0.022(0.010) &0.672(0.006) &0.296(0.000)\\
covariates& $\#S$& 200&20.942(33.596) &15.838(32.890) &18.380(33.082)\\
i.i.d&      & 400&23.610(29.100) &25.170(28.908) &26.438(28.294)\\
generated& PCD& 200 & 0.851 & 0.731 & 0.768\\
& & 400  & 0.918 & 0.820 & 0.858\\
& $\Mean Y^\star(\hat{d})$ &200 &1.556(0.059) &0.752(0.151) &2.005(0.110)\\
& &400 &1.620(0.025) &0.904(0.062) &2.135(0.044)\\
& $\Mean Y^\star(d^{\tiny{opt}})$ & &1.645(0.015) &1.020(0.013) &2.207(0.014)\\ \hline
large & $L_2$ loss & 200 &1.379(1.320) &1.752(1.275) &1.634(1.292)\\
magnitudes&    &400 &0.640(0.725) &0.947(0.720) &0.836(0.727)\\
for $d_{n\beta}$& FN &200&0.194(0.116) &0.330(0.114) &0.232(0.136)\\
with&    &400  &0.000(0.000) &0.014(0.000) &0.002(0.004)\\
covariates& $\#S$& 200&19.962(32.308) &20.864(32.902) &20.738(32.830)\\
i.i.d&      & 400&16.130(28.604) &21.446(27.798) &19.484(28.042)\\
generated& PCD& 200 &0.887 &0.862 &0.876\\
& & 400  &0.950 &0.925 &0.933\\
& $\Mean Y^\star(\hat{d})$ &200 &2.173(0.131) &1.499(0.148) &2.716(0.122)\\
& &400 &2.283(0.029) &1.659(0.037) &2.8135(0.031)\\
& $\Mean Y^\star(d^{\tiny{opt}})$ & &2.285(0.023) &1.659(0.021) &2.848(0.021)\\ \hline
moderate & $L_2$ loss & 200 &0.938(1.382) &1.450(1.387) &1.358(1.399)\\
magnitudes&    &400 &0.490(0.750) &1.105(0.740) &0.841(0.748)\\
for $d_{n\beta}$& FN &200&0.850(0.186) &3.042(0.232) &2.454(0.264)\\
with&    &400  &0.064(0.010) &1.340(0.006) &0.550(0.004)\\
covariates& $\#S$& 200&32.860(22.370) &32.662(13.424) &32.728(17.042)\\
following&      & 400&24.956(28.630) &25.408(28.802) &28.700(29.504)\\
an AR$(1)$& PCD& 200 & 0.827 & 0.669 & 0.710\\
structure& & 400  & 0.916 & 0.776 & 0.835\\
& $\Mean Y^\star(\hat{d})$ &200 &1.459(0.083) &0.612(0.158) &1.859(0.143)\\
& &400 &1.542(0.026) &0.784(0.090) &2.043(0.048)\\
& $\Mean Y^\star(d^{\tiny{opt}})$ & &1.565(0.015) &0.940(0.013) &2.128(0.012)\\ \hline
large & $L_2$ loss & 200 &1.170(1.363) &1.821(1.400) &1.653(1.412)\\
magnitudes&    &400 &0.564(0.725) &0.930(0.747) &0.774(0.757)\\
for $d_{n\beta}$& FN &200&0.148(0.200) &0.580(0.238) &0.442(0.218)\\
with&    &400  &0.000(0.002) &0.000(0.006) &0.004(0.006)\\
covariates& $\#S$& 200&21.776(32.946) &23.338(32.534) &22.994(32.732)\\
following&      & 400&14.958(27.786) &23.328(28.228) &20.330(29.064)\\
an AR$(1)$& PCD& 200 &0.897 &0.841 &0.860\\
structure& & 400  &0.952 &0.920 &0.934\\
& $\Mean Y^\star(\hat{d})$ &200 &2.027(0.107) &1.281(0.204) &2.514(0.174)\\
& &400 &2.097(0.023) &1.444(0.038) &2.644(0.031)\\
& $\Mean Y^\star(d^{\tiny{opt}})$ & &2.111(0.021) &1.486(0.020) &2.672(0.019)\\ \hline
\end{tabular}
\end{table}

Table 1 summarizes the results. The proposed estimators perform well when covariates are independent and the magnitude of half signal is large. When sample size is small, about $2.3$ and $1.7$ important variables in $X_{1\beta}$ in Models II and III are missed on average. However, the selection consistency can be observed either as the sample size increases or as the strength of the signal of $\beta$ increases, as expected. Results on Model I are generally better than those on Model II or III, due to a simple linear form for the baseline function. The value function of the estimated optimal treatment regime slightly increases in all three models as the sample size gets larger, approaching the optimal value function.

The correlation structure of covariates also has influence on the performance of the proposed estimators. When an AR$(1)$ structure is chosen, the estimator tends to miss more important variables and produce larger $L_2$ loss. The second model seems to be the most challenging case. When $n=200$ and the signal is relatively weak, more than three important variables are missed on average. The PCD is only about $66\%$ in this scenario. When the sample size or the magnitude of the signals increase, the performance of the estimators demonstrate similar trend as those obtained for the i.i.d. case.

\subsection{Sensitivity analysis} We conduct some sensitivity analysis when the propensity score model is misspecified. Instead of fitting a penalized logistic
regression model, we estimated the propensity score using sample proportions, i.e., $\hat{\pi}_i=\sum_{i=1}^n A_i/n$. The results are summarized in Table 2.

\begin{table}[htbp]
	\caption{$L_2$, $\#S$, FN, PCD, average mean and standard deviation (in parentheses) of $Y^\star(\hat{d})$ and $Y^\star(d^{\tiny{opt}})$ over 500 simulations for all three models where $n=200$, $400$ and the propensity score model is misspecified}
	\begin{tabular}{cccccc}\hline
		moderate & $L_2$ loss & 200 &0.586 &1.707 &0.960\\
		magnitudes&    &400 &0.230 &0.618 &0.465\\
		for $d_{n\beta}$& FN &200	&0.006 &0.876 &0.396\\
		with&    &400   &0.000 &0.048 &0.002\\
		covariates& $\#S$& 200&31.874 &28.434 &33.224\\
		i.i.d&      & 400&16.810 &34.292 &30.918\\
		generated& PCD& 200 & 0.918 & 0.804 & 0.834\\
		& & 400  & 0.967 & 0.888 & 0.926\\
		& $\Mean Y^\star(\hat{d})$ &200 &1.621(0.021) &0.874(0.083) &2.107(0.046)\\
		& &400 &1.641(0.015) &0.975(0.028) &2.187(0.018)\\
		& $\Mean Y^\star(d^{\tiny{opt}})$ & &1.644(0.015) &1.020(0.013) &2.206(0.013)\\ \hline
		large & $L_2$ loss & 200 &1.091 &1.025 &0.963\\
		magnitudes&    &400 &0.463 &0.424 &0.427\\
		for $d_{n\beta}$ & FN &200&0.000 &0.000 &0.000\\
		with&    &400  &0.000 &0.000 &0.00\\
		covariates& $\#S$& 200&23.526 &28.942 &26.232\\
		i.i.d&      & 400&9.594 &16.106 &13.110\\
		generated& PCD& 200 &0.948 &0.927 &0.936\\
		& & 400  &0.984 &0.968 &0.978\\
		& $\Mean Y^\star(\hat{d})$ &200 &2.264(0.028) &1.620(0.036) &2.819(0.033)\\
		& &400 &2.283(0.022) &1.655(0.022) &2.844(0.023)\\
		& $\Mean Y^\star(d^{\tiny{opt}})$ & &2.284(0.023) &1.661(0.022) &2.847(0.021)\\ \hline
		moderate & $L_2$ loss & 200 &0.574 &1.251 &1.066\\
		magnitudes&    &400 &0.230 &0.691 &0.476\\
		for $d_{n\beta}$& FN &200	&0.026 &1.786 &0.936\\
		with&    &400   &0.000 &0.154 &0.004\\
		covariates& $\#S$& 200&33.890 &25.170 &32.092\\
		following&      & 400&17.708 &40.028 &34.708\\
		an AR$(1)$& PCD& 200 & 0.909 & 0.751 & 0.796\\
		structure& & 400  & 0.962 & 0.862 & 0.913\\
		& $\Mean Y^\star(\hat{d})$ &200 &1.540(0.023) &0.732(0.111) &1.994(0.079)\\
		& &400 &1.561(0.016) &0.882(0.034) &2.107(0.018)\\
		& $\Mean Y^\star(d^{\tiny{opt}})$ & &1.565(0.015) &0.940(0.013) &2.128(0.012)\\ \hline
		large & $L_2$ loss & 200 &1.031 &0.993 &0.930\\
		magnitudes&    &400 &0.399 &0.408 &0.405\\
		for $d_{n\beta}$& FN &200&0.000 &0.000 &0.002\\
		with&    &400  &0.000 &0.000 &0.00\\
		covariates& $\#S$& 200&18.746 &32.254 &27.026\\
		following&      & 400&10.214 &15.532 &12.966\\
		an AR$(1)$& PCD& 200 &0.955 &0.918 &0.932\\
		structure& & 400  &0.983 &0.964 &0.972\\
		& $\Mean Y^\star(\hat{d})$ &200 &2.096(0.023) &1.443(0.041) &2.644(0.031)\\
		& &400 &2.109(0.019) &1.478(0.020) &2.667(0.020)\\
		& $\Mean Y^\star(d^{\tiny{opt}})$ & &2.111(0.021) &1.486(0.020) &2.672(0.019)\\ \hline
	\end{tabular}
\end{table}

The estimators perform comparably to the cases when we the propensity score model is correctly specified, which 
suggests that the proposed estimators may not be sensitive to misspecified propensity score models. 

\section{Real data example}
We further examine our method on the data set from the NIMH-funded STAR*D study, where $4041$ patients in total with nonpsychotic major depressive disorder (MDD) were participated. The aim of the study was to determine the effectiveness of different treatments for those people who have not responded to initial medication treatment. Patients first received citalopram (CIT), an SSRI medication. After $8$-$12$ weeks, three more levels of treatments were offered to participants whose first treatment didn't give an acceptable response. Available treatment strategies at Level two includes sertraline (SER), venlafaxine (VEN), bupropion (BUP) and cognitive therapy (CT) and augmenting CIT which combines CIT with one more treatment. Two Level 2A switch options with VEN and BUP treatments were provided for patients receiving CT without sufficient improvement. Four treatments were available at Level 3 for participants without anticipated response, including medication switch to mirtazapine (MIRT), nortriptyline (NTP), and medication augmentation with either lithium (Li) and thyroid hormone (THY). Finally, treatment with tranylcypromine (TCP) or a combination of mirtazapine and venlafaxine (MIRT+VEN) were provided at Level 4 for those without sufficient improvement at Level 3.
	
Here, we compare the treatment strategies for BUP (coded as $1$) and SER ($0$) at Level 2. This gives all together $319$ patients with complete records of covariates and response. Among them, $153$ were treated with BUP and $166$ with SER. The outcome was measured by the 16-item Quick Inventory of Depressive Symptomatology-Clinician-Rated (QIDS-C16), which indicated the severity of patient's depressive symptom. All baseline variables at Level 1 and intermediate outcomes at Level 2 are included in the study, yielding $305$ covariates in total for each patient. A linear model with all $305$ covariates are fitted for the main-effect model. Both LASSO and SCAD penalty functions are used. LASSO selects $4$ important variables, including IMPWR (indicting whether patients thought they have special powers), URDIF (difficulty in urination), URPN (painful urination) and QCCURRATE (QIDS-C score changing rates at Level 1), yielding the following optimal treatment regime:
\begin{equation*}
I(0.59 \mbox{IMPWR}+0.29 \mbox{URDIF} +1.33 \mbox{URPN} + 0.66 \mbox{QCCURRATE}>0).
\end{equation*}

Apart from these $4$, The SCAD penalty selects another $4$ variables, including  HAWAI (Native Hawaiian/other Pacific Islander), SKRSH (rash skin or not), NVTRM (CNS: Tremors) and URNONE (no symptoms in patients' urination category). Optimal treatment regime using SCAD penalty is given as
\begin{eqnarray*}
I(-0.01\mbox{HAWAI}+1.33\mbox{IMPWR}-0.19\mbox{SKRSH}+0.30\mbox{NVTRM}\\
+0.79\mbox{URDIF}+1.66\mbox{URPN}-0.12\mbox{URNONE}+0.64\mbox{QCCURRATE}>0).
\end{eqnarray*}

\section{Conclusion}
In this article, we propose a two-step estimator for estimating the optimal treatment strategy, which selects variables and estimates parameters simultaneously in both propensity score and outcome regression models using penalized regression. Our methodology can handle data set whose dimensionality is allowed to grow exponentially fast compared to the sample size. Oracle properties of the estimators are given. Variable selection is also involved in the misspecified model and new mathematical techniques are developed to study the estimator's properties in a general form of optimization. The estimator is shown to be more efficient when the misspecified working model is ``closer'' to the conditional mean of the response, although our approach does not require correct specification of the baseline function. Numerical results demonstrate that the proposed estimator enjoys model selection consistency and has overall satisfactory performance.

In the case when there are multiple local solutions of our objective functions (\ref{beta}), (\ref{alpha}) or (\ref{theta}), although our asymptotic theory only suggests the existence of a local minimum possessing the oracle property, it is worth mentioning that we can actually identify the desired oracle estimator using existing algorithms \citep[see][]{fan2014, wang2013}. Theoretical properties can be established in a similar fashion.

The current framework is focused on point exposure study. It will be interesting and practically useful to extend our results to dynamic treatment regimes. 
Significant efforts are needed to handle model misspecification in multiple stages. This is beyond the scope of the current paper and is an interesting future research topic.

\newpage
\appendix
\section{Technical conditions}
We give some regularity conditions on the design matrix and constraints on the penalty function and regularization parameters here. Condition 2 and 5 serve to guarantee weak oracle property of the estimator while Condition 2$^\prime$ and 5$^\prime$ ensure the oracle property.
\subsection{The design matrix}\label{A.1}
\begin{cond}\label{condregudesignmatrix}
	The design matrix $X$ satisfies
	\begin{eqnarray}\label{x1x1}
	&&||(X_{1\alpha}^T \Delta X_{1\alpha})^{-1}||_\infty = O(\frac{b_\alpha}{n}),\quad ||(X_{1\beta}^T \Delta X_{1\beta})^{-1}||_\infty = O(\frac{b_\beta}{n}),\\ \label{x2x1}
	&&||X_{2\alpha}^T \Delta X_{1\alpha} (X_{1\alpha}^T \Delta X_{1\alpha})^{-1}||_\infty \le \min\left\{C \frac{\rho^{'}_1(0+)}{\rho^{'}_1(d_{n\alpha})}, O(n^{a_1})\right\},\\ \label{x2bx1b}
	&&||X_{2\beta}^T \Delta X_{1\beta} (X_{1\beta}^T \Delta X_{1\beta})^{-1}||_\infty \le \min\left\{C \frac{\rho^{'}_2(0+)}{\rho^{'}_2(d_{n\beta})}, O(n^{a_2})\right\},\\
	\label{lamx1a}&&\max_{j=1}^p \lambda_{\max} \{X_{1\alpha}^T \diag (|x^j|)X_{1\alpha}\} = O(n),\\
	\label{lamx1b}&&\max_{j=1}^p \lambda_{\max} \{X_{1\beta}^T \diag (|x^j|)X_{1\beta} \}=O(n),
	\end{eqnarray}
	and if the response is unbounded,
	\begin{eqnarray}\label{maxentry}
	\max_{j=1}^p (||x^j||_\infty) = o(n^{d_\beta}/\sqrt{\log n}), \qquad\qquad\qquad\qquad~~
	\end{eqnarray}
	where $a_1$ and $a_2$ are constants in $\left[0, \frac{1}{2}\right]$, $C \in (0, 1)$, $d_\beta$ a positive constant that satisfies $\log(p)=O(n^{1-2d_\beta})$. $b_\alpha$ and $b_\beta$ are two sequences that are allowed to diverge. $\Delta$ is a diagonal matrix with the $i$th diagonal element $\pi_i(1-\pi_i)$.
\end{cond}


Condition (\ref{x1x1}), (\ref{x2x1}), (\ref{x2bx1b}), (\ref{lamx1a}) and (\ref{lamx1b}) are the regularity conditions on the design matrix. 
They share similar purpose as Condition 2 in \cite{fan2011}.
Condition (\ref{maxentry}) controls the order of magnitude of the largest element in the design matrix. We require the propensity score bounded away from $0$ and $1$ to make $X_{1\alpha}^T \Delta X_{1\alpha}$ and $X_{1\beta}^T \Delta X_{1\beta}$ non-degenerate. As commented in \cite{fan2011}, these constraints are easy to be satisfied. Since the intercept is contained in the design matrix, condition (\ref{lamx1a}) and (\ref{lamx1b}) imply that $\lambda_{\max}(X_{1\alpha}^T X_{1\alpha})=\lambda_{\max}(X_{1\beta}^T X_{1\beta})=O(n)$. 
\begin{cond}\label{condregudesignmatrix2}
The design matrix shall satisfy
\begin{eqnarray}\label{lamminx}
&&\displaystyle\lambda_{\min} (X_{1\alpha}^T \Delta X_{1\alpha}) \ge b_1 n, \quad \lambda_{\min}(X_{1\beta}^T \Delta X_{1\beta}) \ge b_2 n, \qquad\qquad \\
&&\label{x2x12infty}||X_{2\alpha}^T \Delta X_{1\alpha}||_{2,\infty} = O(n),\quad ||X_{2\beta}^T \Delta X_{1\beta}||_{2,\infty} = O(n), \qquad\qquad \\
&&\label{xinfinity}\displaystyle\max_{j=1}^p ||x^j||_\infty = o(n^{d_\beta}/\sqrt{\log n}),
\end{eqnarray}
and conditions \eqref{lamx1a}, \eqref{lamx1b} in Condition \ref{condregudesignmatrix}, where $b_1, b_2$ are some positive constants, and $\displaystyle||B||_{2,\infty} = \sup_{||v||=1} ||Bv||_\infty$.
\end{cond}

Conditions (\ref{lamminx}) and (\ref{x2x12infty}) are generally stronger than (\ref{x1x1}), (\ref{x2x1}) and (\ref{x2bx1b}). Condition (\ref{lamminx}) satisfies when the sequence of $b_\alpha$ and $b_\beta$ in (\ref{x1x1}) are bounded. In addition, if $s_\alpha$ and $s_\beta$ are bounded, (\ref{lamminx}) implies (\ref{x1x1}) with $b_\alpha=b_\beta=1$ and (\ref{balpha}) also holds. 

Similar to (\ref{x2x1}), (\ref{x2x12infty}) requires that the correlations between the covariates in $X_{1\alpha}$ and those in $X_{2\alpha}$, as well as the correlations between covariates in $X_{1\beta}$ and those in $X_{2\beta}$, are not too strong, since each element in the matrix corresponds to the inner product of each important and unimportant variables adjusted by the weights $\pi_i(1-\pi_i)$. When (\ref{lamminx}) and (\ref{x2x12infty}) hold, the right-hand order $O(n^{a_1})$ and $O(n^{a_2})$ in (\ref{x2x1}) and (\ref{x2bx1b}) hold as long as $a_1\ge\sqrt{s_\alpha}$, $a_2\ge\sqrt{s_\beta}$.
\subsection{Penalty and regularization}\label{A.2}
Let
\begin{eqnarray*}
	d_{n\alpha} &=& \frac{1}{2} \min_j \left\{|\alpha_{0,j}|:\alpha_{0,j}\ne 0\right\}, \quad
	d_{n\beta} = \frac{1}{2} \min_j \left\{|\beta_{0,j}|:\beta_{0,j}\ne 0|\right\},
\end{eqnarray*}
denote the half minimal signal of $\alpha_0$ and $\beta_0$. We choose the regularization parameter $\lambda_{1n}$, $\lambda_{2n}$, $\lambda_{3n}$ and introduce the following condition.

\begin{cond}\label{condlambda}
Assume that
\begin{eqnarray}\quad
d_{n\alpha} &\ge& n^{-\gamma_\alpha} \log n, \quad d_{n\beta} \ge n^{-\gamma_\beta} \log n, \quad d_{n\theta} \gg n^{-\gamma_\theta}\log n,\\
\label{balpha}b_\alpha &=& o\left\{\min (n^{\frac{1}{2}-\gamma_\alpha} \sqrt{\log n}, n^{\gamma_\alpha}/s_\alpha \log n)\right\}, \\ \label{dbeta}
b_\beta &=& o\left\{\min (n^{\frac{1}{2}-\gamma_\beta} \sqrt{\log n}, n^{2\gamma_\alpha-\gamma_\beta}/s_\alpha \log n, n^{\gamma_\beta}/s_\beta \log n\right\}, \\ 
\label{btheta} b_\theta &=& o\left\{\min (n^{\frac{1}{2}-\gamma_\theta} \sqrt{\log n}, n^{\gamma_\theta}/s_\theta \log n)\right\},
\end{eqnarray}
In addition, take $\lambda_{1n}$, $\lambda_{2n}$, $\lambda_{3n}$ satisfying 
\begin{eqnarray}\label{lambda}
&\lambda_{1n} \rho_1^{'}(d_{n\alpha}) = o\{\min (n^{-\gamma_\alpha} \log n/b_\alpha, n^{-\gamma_\beta} \log n/b_{\alpha\beta})\},\\ \label{lambda2}\quad
&\lambda_{2n} \rho_2^{'}(d_{n\beta}) = o(n^{-\gamma_\beta} \log n/b_\beta),\quad \lambda_{3n} \rho_3^{'}(d_{n\theta}) = o(n^{-\gamma_\theta}\log n/b_\theta),\\ \label{lamn}
&\lambda_{1n}\gg n^{-d_\alpha}\log^2 n,\quad \lambda_{2n} \gg n^{-d_\beta} \log^2 n,\quad \lambda_{3n} \gg n^{-d_\theta} \log^2 n,\\ \label{lam3}
&\lambda_{1n} \kappa_\alpha = o(\tau_\alpha), \quad \lambda_{2n} \kappa_\beta = o(\tau_\beta), \quad \lambda_{3n} \kappa_\theta = o(\tau_\theta),
\end{eqnarray}
where
\begin{eqnarray}
\label{dalpha}&&d_\alpha=\min(\frac{1}{2}+a_1, 2\gamma_\alpha-l_1)-a_1,\\
\label{dbeta0}&& d_\beta = \min(\frac{1}{2}+a_2,2\gamma_\alpha-l_1,2\gamma_\beta - l_2)-a_2,\\
\label{dtheta0}&& d_\theta = \min(\frac{1}{2}+a_3,2\gamma_\theta-l_3)-a_3
\end{eqnarray}
with $s_\alpha = O(n^{l_1})$, $l_1 < \gamma_\alpha$, $s_\beta = O(n^{l_2})$, $l_2 < \min(\gamma_\alpha, \gamma_\beta)$, $s_\theta = O(n^{l_3})$, $l_3 < \gamma_\theta$ and $\displaystyle \kappa_\alpha = \max_{\delta \in H_\alpha} \kappa(\rho_1, \delta)$, $\displaystyle\kappa_\beta = \max_{\delta \in H_\beta} \kappa(\rho_2, \delta)$, $\displaystyle \kappa_\theta = \max_{\theta \in H_\theta} \kappa(\rho_3, \delta)$ with function $\kappa(\rho, v)$ at $v=(v_1, \dots, v_k)^T$ with $||v||_0=k$ as
\begin{eqnarray*}
	\displaystyle \kappa(\rho, v) = \lim_{\epsilon\to 0^{+}} \max_{1\le j\le k} \sup_{t_1<t_2\in(|v_j|-\epsilon, |v_j|+\epsilon)} -\frac{\rho^{'}(t_2)-\rho^{'}(t_1)}{t_2-t_1}.
\end{eqnarray*}
$\displaystyle \tau_\alpha =\lambda_{\min} (\frac{X_{1\alpha}^T \Delta X_{1\alpha}}{n})$, $\displaystyle \tau_\beta = \lambda_{\min} (\frac{X_{1\beta}^T \Delta X_{1\beta}}{n})$, $\displaystyle \tau_\theta =\min_{\delta \in H_\theta}\lambda_{\min}\{\frac{\phi_1(\delta)^T \phi_1(\delta)}{n}\}$.
\end{cond}

These conditions guarantee the existence of estimators. Apart from (\ref{dbeta}), (\ref{lambda}) and (\ref{dbeta0}), other conditions on the penalty and regularization parameter are similar in rationale as in \cite{fan2011}.

\begin{cond}\label{condlambda2}
Assume
\begin{eqnarray}
&&d_{n\alpha} \gg \lambda_{1n} \gg \max\left\{(s_\alpha/n)^{1/2},n^{d_\beta}(\log n)^{1/2}\right\},\\
&&d_{n\beta} \gg \lambda_{2n} \gg \max\left\{(s_\beta/n)^{1/2},n^{d_\beta}(\log n)^{1/2}\right\}, \\
\label{lambdadn}&&\lambda_{1n} \rho_1(d_{n\alpha}) = O(n^{-1/2}), \quad\lambda_{2n} \rho_2(d_{n\beta})=O(n^{-1/2}), \\
\label{kappao1}&&\lambda_{1n} \kappa_\alpha = o(1), \quad\lambda_{2n} \kappa_\beta = o(1),
\end{eqnarray}
where $\kappa_\alpha = \max_{\delta \in H_\alpha} \kappa(\rho_1, \delta)$, $\kappa_\beta = \max_{\delta \in H_\beta} \kappa(\rho_2, \delta)$.
\end{cond}

\section{Proof of Proposition 1}
We proof the case of unbounded variables only. Denote the set of functions $\mathcal{F}$ as $\{A^j(h_1)-A^j(h_2)\}^T e$, $\forall j=1,\dots,J, h_1,h_2 \in H$, by the Symmetrization Lemma \citep[Lemma 2.3.1,][]{van1996}, we have
\begin{eqnarray*}
&\displaystyle \Mean \sup_{h_1,h_2 \in H}||\{A(h_1)-A(h_2)\}^T z||_\infty \le \Mean \max_{j=1}^J \sup_{h_1,h_2 \in H}|\{A^j(h_1)-A^j(h_2)\}^T z|\\
&\displaystyle \le \Mean_e \Mean_\epsilon 2\max_{j=1}^J \sup_{h_1,h_2 \in H}|\sum_i \{a_{ij}(h_1)-a_{ij}(h_2)\} z_i \epsilon_i|,
\end{eqnarray*}
where $\epsilon_i$'s are independent Rademacher variables.

If we can show
\begin{eqnarray}\label{prop3eq2}
&\displaystyle \Mean_\epsilon \max_{j=1}^J \sup_{h_1,h_2 \in H}|\sum_i \{a_{ij}(h_1)-a_{ij}(h_2)\} z_i \epsilon_i|\\ \nonumber
&\displaystyle \le \max_{i=1}^n |z_i|O\{\delta \sqrt{S\log (J)} \max_{j=1,\dots,J} \sup_{h\in H} ||b_j(h)||_2\},
\end{eqnarray}
(\ref{prop1eq1}) follows by $\displaystyle \Mean \max_{i=1}^n |z_i| \le ||\max_{i=1}^n |z_i| ||_{\psi_1} \le \log n ||z_i||_{\psi_1}$.

It follows from Lemma 2.2.2 in \cite{van1996} and $\Mean |X|\le \sqrt{\Mean |X|^2} \le ||X||_{\psi_2}$ that
\begin{eqnarray*}
 \displaystyle&&\Mean_\epsilon \max_{j=1}^J \sup_{h_1,h_2 \in H}|\sum_i \{a_{ij}(h_1)-a_{ij}(h_2)\} z_i \epsilon_i|\\
\displaystyle\le&&||\max_{j=1}^J \sup_{h_1,h_2 \in H}\sum_i \{a_{ij}(h_1)-a_{ij}(h_2)\} z_i \epsilon_i||_{\psi_2}\\
\displaystyle\le&&\sqrt{\log J}\max_{j=1}^J||\sup_{h_1,h_2 \in H}\sum_i \{a_{ij}(h_1)-a_{ij}(h_2)\} z_i \epsilon_i||_{\psi_2},
\end{eqnarray*}
therefore it suffices to show
\begin{eqnarray}\label{prop3eq3}
&\displaystyle \max_{j=1}^J||\sup_{h_1,h_2 \in H}\sum_i \{a_{ij}(h_1)-a_{ij}(h_2)\} z_i \epsilon_i||_{\psi_2}\\ \nonumber
&\displaystyle =O(\delta\sqrt{S}\max_{i=1}^n |z_i| \max_{j=1,\dots,J} \sum_{s=1}^S \sup_{h\in H} ||b_j^s(h)||_2).
\end{eqnarray}

For fixed $z_i$, we prove the stochastic process $\sum_{i} a_{ij}(h) z_i \epsilon_i$ indexed by $h$ is sub-gaussian with the semi-metric $$\displaystyle d(h_1,h_2)=\max_{i=1}^n |z_i| \max_{j=1,\dots,J} \sup_{h\in H} ||b_j(h)||_2 ||h_1-h_2||_\infty.$$

By Lemma \ref{lemmac1} in Appendix C, it suffices to show
\begin{eqnarray}\label{prop3eq4}
\displaystyle \max_j \sup_{h_1,h_2 \in H} \sqrt{\sum_i \{a_{ij}(h_1) - a_{ij}(h_2)\}^2 z_i^2}\le d(h_1,h_2).
\end{eqnarray}

It follows from mean-value theorem that
\begin{eqnarray*}
&&\displaystyle \max_j \sup_{h_1,h_2 \in H} \sqrt{\sum_i \{a_{ij}(h_1) - a_{ij}(h_2)\}^2 z_i^2}\\
\displaystyle &\le& \max_{i=1}^n |z_i| \max_{j} \sup_{h_1,h_2 \in H} \sqrt{\sum_i \{a_{ij}(h_1) - a_{ij}(h_2)\}^2}\\
\displaystyle &\le& \max_{i=1}^n |z_i| \max_{j} \sup_{h_1,h_2 \in H} \sqrt{\sum_i \{\sum_{s=1}^S b_{ij}(\tilde{h}_s) (h_{1s}-h_{2s})\}^2}
\end{eqnarray*}
where $\tilde{h}_s$, $\forall s=1,\dots,S$ lies between the line segment of $h_1$ and $h_2$. Together with the fact that
\begin{eqnarray*}
&&\displaystyle \sum_i \{\sum_{s=1}^S b_{ij}(h_s) (h_{1s}-h_{2s})\}^2\\
\le&& \displaystyle \sum_i \{\sum_s b_{ij}^2(h_s) (h_{1s}-h_{2s})^2 + \sum_{s_1\neq s_2}b_{ij}(h_{1s_1}) b_{ij}(h_{s_2})(h_{1s_1}-h_{s_2})^2\}\\
\le&& \displaystyle ||h_1-h_2||_\infty^2\{\sum_i\sum_s b^2_{ij}(h_s)+\sum_{s_1\neq s_2}\sum_i b_{ij}(h_{s_1})b_{ij}(h_{s_2})\}\\
\le&& \displaystyle ||h_1-h_2||_\infty^2\{\sum_s\sum_i b^2_{ij}(h_s)+\sum_{s_1\neq s_2}\sqrt{\sum_i b^2_{ij}(h_{s_1})} \sqrt{\sum_i b^2_{ij}(h_{s_1})}\}\\
\le&& \displaystyle ||h_1-h_2||_\infty^2\{\sum_s \sqrt{\sum_i b^2_{ij}(h_s)}\}^2,
\end{eqnarray*}
we have
\begin{eqnarray*}
&&\displaystyle \max_j \sup_{h_1,h_2 \in H} \sqrt{\sum_i \{a_{ij}(h_1) - a_{ij}(h_2)\}^2 z_i^2}\\
&\le&\displaystyle \max_i |z_i| \max_j \sup_{h_1,h_2\in H}||h_1-h_2||_\infty\{\sum_s \sqrt{\sum_i b^2_{ij}(\tilde{h}_s)}\}\\
&\le&\displaystyle \delta \max_i |z_i| \max_j \{\sum_s \sup_{h\in H} \sqrt{\sum_i b_{ij}^2(h)}\},
\end{eqnarray*}
which implies (\ref{prop3eq4}).

By some arguments for sub-gaussian process in the proof of Corollary 2.2.8 in \cite{van1996}, we have
\begin{eqnarray*}
&\displaystyle ||\sup_{h_1,h_2 \in H}\sum_i \{a_{ij}(h_1)-a_{ij}(h_2)\} z_i \epsilon_i||_{\psi_2}\\
&\le\displaystyle K\int_0^{\delta_n}\sqrt{\log D(\epsilon, d)}d\epsilon\le K\int_0^{\delta_n}\sqrt{\log N(\frac{\epsilon}{2}, d)}d\epsilon
\end{eqnarray*}
where $\displaystyle \delta_n = \delta \max_{i=1}^n |z_i| \max_{j} \sum_{s=1}^S\sup_{h_1,h_2 \in H} ||b^s_j(h)||_2$. Since $d(h_1, h_2)$ is proportional to $||h_1-h_2||_\infty$, we have
\begin{eqnarray*}
N(\epsilon, d)\le M(\frac{\delta_n}{\epsilon})^S,
\end{eqnarray*}
with $M$ some constant independent of $\epsilon$ and $j$, thus (\ref{prop3eq3}) is satisfied. This completes the proof for (\ref{prop1eq1}). To show (\ref{prop1eq2}), define
\begin{eqnarray*}
\beta_n(t)=\inf_{\substack{j \in [1,\dots,J]\\h_1,h_2\in H}} \prob\left(|\sum_i [a_{ij}(h_1)-a_{ij}(h_2)]z_i|<\frac{t}{2}\right).
\end{eqnarray*}

It follows by Chebyshev's inequality and the condition $\max_i \Mean |z_i|^2=v_0^2$ that for any $t\ge 3\delta d_nv_0$,
\begin{eqnarray*}
 \beta_n(t)\ge 1-\frac{4v_0^2\sup_{h_1,h_2\in H}\sum_{i}[a_{ij}(h_1)-a_{ij}(h_2)]^2}{t^2}\ge 1-\frac{1}{2}=\frac{1}{2},
\end{eqnarray*}
where the last inequality follows due to that
\begin{eqnarray*}
\sup_{h_1,h_2\in H}\sum_{i}[a_{ij}(h_1)-a_{ij}(h_2)]^2\le \delta^2 d_n^2,
\end{eqnarray*}
using similar arguments in showing (\ref{prop3eq4}). Now it follows by Lemma 2.3.7 in \cite{van1996} that
\begin{eqnarray}\nonumber
&&\prob\left(\sup_{\substack{j=1,\dots,J\\h_1,h_2\in H}}|\sum_i [a_{ij}(h_1)-a_{ij}(h_2)]z_i|>t\right)\\ \label{prop1eq3}
&\le& 4\prob\left(\sup_{\substack{j=1,\dots,J\\h_1,h_2\in H}}|\sum_i [a_{ij}(h_1)-a_{ij}(h_2)]z_i\epsilon_i|>t\right),
\end{eqnarray}
where $\epsilon_1,\dots,\epsilon_n$ are i.i.d Rademacher variables. Define $\mathcal{A}$ to be the event in (\ref{prop1eq3}), $\mathcal{B}$ the event
\begin{eqnarray*}
\max_{i} |z_i|\le \omega\log^2 n,
\end{eqnarray*}
we can further bound (\ref{prop1eq3}) from above by
\begin{eqnarray*}
4\prob(\mathcal{A}\cap\mathcal{B})+4\prob(\mathcal{B}^c).
\end{eqnarray*}

We first show $\prob(\mathcal{B}^c)\le 2/n$. By the definition of the Orlicz norm $||\cdot||_{\psi_1}$ and Markov's inequality,
\begin{eqnarray}\label{prop1eq4}
\prob(\mathcal{B}^c)\le \frac{\exp(||\max_i |z_i|||_{\psi_1}/\omega\log n)}{\exp(\omega \log^2n/\omega\log n)}\le \frac{2}{\exp(\log n)}=\frac{2}{n},
\end{eqnarray}
since $||\max_i |z_i|||_{\psi_1}\le \log n\max_i ||z_i||_{\psi_i}\le \omega\log n$. Besides, it follows by Bonferroni's inequality that
\begin{eqnarray*}
\prob(\mathcal{A}\cap\mathcal{B})\le J\max_j \prob\left(\sup_{h_1,h_2\in H}|\sum_i [a_{ij}(h_1)-a_{ij}(h_2)]z_i\epsilon_i|>t, \max_{i} |z_i|\le \omega\log^2 n\right).
\end{eqnarray*}

On the event $\mathcal{B}$, conditional on $z_i$, denote
\begin{eqnarray*}
Y_j=\sup_{h_1,h_2\in H}|\sum_i [a_{ij}(h_1)-a_{ij}(h_2)]z_i\epsilon_i|,
\end{eqnarray*}
it follows by (\ref{prop3eq3}) that
\begin{eqnarray*}
\max_j ||Y_j||_{\psi_2}\le c_0 \delta \max_i |z_i| d_n\le c_0\delta d_n \omega\log^2 n,
\end{eqnarray*}
for some constant $c_0$. Hence, on event $\mathcal{B}$, conditional on $z_i$'s,
\begin{eqnarray*}
\max_j \prob\left(|Y_j|\ge t\right)\le  \max_j \frac{\exp([|Y_j|/c_0 \delta d_n \omega\log^2 n]^2)}{\exp([t/c_0\delta d_n \omega\log^2 n]^2)}\le \exp\left(-\frac{2t^2}{c_0^2\delta^2 d_n^2 \omega^2\log^4 n}\right).
\end{eqnarray*}

Note that the right-hand side is independent of $z_i$, thus using the law of total expectation, we have
\begin{eqnarray*}
\max_j \prob(\mathcal{A}\cap \mathcal{B})\le J\exp\left(-\frac{2t^2}{c_0^2\delta^2 d_n^2 \omega^2\log^4 n}\right),
\end{eqnarray*}
which together with (\ref{prop1eq3}) and (\ref{prop1eq4}) gives the result.

\newpage
\section{Proof of Theorem 1}
Before proving Theorem 1, we state the following lemmas. The proof of Lemma 2 is given in Appendix G.
\begin{lemma}\label{lemmac1}
Let $z=(z_1,\dots,z_n)^T$ be an $n$-dimensional independent random response vector with mean $0$ and $a \in \mathbb{R}^n$.
	\begin{enumerate}
		\item [(a)] If $z_1, \dots, z_n$ are bounded in $[c,d]$, then for any $\epsilon \in (0, \infty)$,
		\begin{eqnarray*}
			\prob (|a^T z|>\epsilon) \le 2 \exp\left(-\frac{\epsilon^2}{2||a||_2^2 (d-c)^2}\right).
		\end{eqnarray*}
		\item [(b)] If $z_1, \dots, z_n$ satisfy $\max_i ||z_i||_{\psi_1}\le \omega$, then for any $\epsilon \in (0, \infty)$,
		\begin{eqnarray*}
			\prob (|a^T z|>\epsilon) \le 2 \exp\left(-\frac{1}{2}\frac{\epsilon^2}{2||a||_2^2 \omega^2 + ||a||\epsilon \omega}\right).
		\end{eqnarray*}
	\end{enumerate}
\end{lemma}

\begin{lemma}
Define $\varepsilon=\cup_{k=1}^{16}\varepsilon_k$, where $\varepsilon_k$ is defined in Appendix G, under conditions in Theorem 1, we have $\prob(\varepsilon)\ge 1-\bar{c}/(n+p+q)$ for some $\bar{c}>0$.
\end{lemma}

\begin{note}
Let $Z = \diag(A-\pi) X$, $\hat{Z} = \diag(A-\hat{\pi}) X$, and $Z_1$, $Z_2$, $\hat{Z}_1$, $\hat{Z}_2$ the sub-matrices of $Z$ and $\hat{Z}$, formed by components in supp$(\beta_0)$ and its complement, respectively. Define
\begin{eqnarray*}
	&&\xi_1 = \hat{Z}^T e, \quad \xi_2 = Z^T (\mu-\Phi), \quad \xi_3 = \phi^T (e-Z\beta_0),\\
	&&\xi_4 = Z^T \diag(X \beta_0) \Delta X_{1\alpha},\quad \xi_5 = X^T [\diag\left\{(A - \pi)\circ(A-\pi)\right\}-\Delta] X_{1\beta},\\
	&&\xi_6(\delta) = Z^T \{\Phi - \Phi(\delta)\},\quad\xi_7(\delta)=\{\phi(\delta)-\phi\}^T (e-Z \beta_0),
\end{eqnarray*}
and $M_{\beta} = \mbox{supp}(\beta_0)$, $M_{\beta}^\prime = \mbox{supp}(\theta^\star)$ and their complements $M_{\beta}^c$, $M_{\theta}^{c\prime}$ respectively. For a given vector $\psi$ or a matrix $\Psi$, $\psi_{\mathbb{M}}$ stands for the sub-vector of $\psi$ formed by components in $\mathbb{M}$, $\Psi_{\mathbb{M}}$ the sub-matrix of $\Psi$ by rows in $\mathbb{M}$. Besides, the superscript $\Psi^j$ is used to refer to the vector which is the $j$th column of matrix $\Psi$ while the subscript $\Psi_i$ stands for the $i$th row of $\Psi$.
\end{note}

\textit{Proof of theorem 1.} We break the proof into three steps. Based on Theorem 1 in \cite{fan2011}, it suffices to prove the existence of $\hat{\beta}_1$, $\hat{\theta}_1$ inside the hypercube
\begin{eqnarray*}
\aleph=\left\{(\delta_\beta^T, \delta_\theta^T)^T: ||\delta_\beta - \beta_1||_\infty = n^{-\gamma_\beta}\log n, ||\delta_\theta - \theta_1^\star||_\infty = Kn^{-\gamma_\theta}\log n \right\}
\end{eqnarray*}
with $K$ a large constant, conditional on the event $\varepsilon$, satisfying
\begin{eqnarray}
\label{betaeqcon}&&\hat{Z}_1^T \{Y-\Phi(\hat{\theta})-\hat{Z}\hat{\beta}\}=n\lambda_{2n} \bar{\rho}_2(\hat{\beta}_1),\\
\label{thetaeqcon}&&\hat{\phi}_1^T \{Y-\Phi(\hat{\theta})\}=n\lambda_{3n} \bar{\rho}_3(\hat{\theta}_1),\\
\label{betaineqcon}&&||\hat{Z}_2^T \{Y-\Phi(\hat{\theta})-\hat{Z}\hat{\beta}\}||_\infty < n\lambda_{2n} \rho^{'}_2(0+), \\
\label{thetaineqcon}&&||\hat{\phi}_2^T \{Y-\Phi(\hat{\theta})\}||_\infty < n\lambda_{3n} \rho^{'}_3(0+), \\
\label{betasdcon}&&\lambda_{\min}(\hat{Z}_1^T \hat{Z}_1) > n\lambda_{2n}\kappa(\rho_2, \hat{\beta}_1),\\
\label{thetasdcon}&&\lambda_{\min}(\hat{\phi}_1^T \hat{\phi}_1) > n\lambda_{3n}\kappa(\rho_3, \hat{\theta}_1).
\end{eqnarray}


\textit{Step 1}. We first show the existence of a solution to equations (\ref{betaeqcon}) and (\ref{thetaeqcon}) inside $\aleph$ for sufficiently large $n$. For any $\delta = (\delta_1, \dots, \delta_{s_\beta+s_\theta})^T \in \aleph$, since $d_{n\beta} \ge n^{-\gamma_\beta}\log n$, $d_{n\theta} \gg n^{-\gamma_\theta}\log n$, we have
\begin{eqnarray*}
\min_{j=1}^{s_\beta} |\delta_j| \ge \min |\beta_{0,j}| - d_{n\beta} = d_{n\beta}, \quad \min_{j=1}^{s_\theta} |\delta_{j+s_\beta}| \ge \min |\theta_j^\star| - d_{n\theta} = d_{n\theta}
\end{eqnarray*}
and $\mbox{sgn}(\delta_\beta) = \mbox{sgn}(\beta_1)$, $\mbox{sgn}(\delta_\theta) = \mbox{sgn}(\theta_1^\star)$. The monotonicity condition of $\rho_2^{'}(t)$, $\rho_3^{'}(t)$ gives
\begin{eqnarray}\label{betapenalty}
||n\lambda_{2n} \bar{\rho}_2(\delta)||_\infty \le n\lambda_{2n} \rho_2^{'}(d_{n\beta}), ||n\lambda_{3n} \bar{\rho}_3(\delta)||_\infty \le n \lambda_{3n} \rho_3^{'}(d_{n\theta}).
\end{eqnarray}

We write the left hand side of (\ref{betaeqcon}) as
\begin{eqnarray}\label{term}
&\hat{Z}_1^T \{Y - \Phi(\delta_\theta) - \hat{Z}_1\delta_\beta\} = \xi_{1M_{\beta}} + \xi_{2M_{\beta}} + (\hat{Z}_1 - Z_1)^T \{\mu-\Phi(\delta_\theta)\}
\\ \nonumber  &+ \hat{Z}_1^T \hat{Z}_1 (\beta_1 - \delta_\beta) + \hat{Z}_1^T (\hat{Z}_1 - Z_1) \beta_1 - Z_1^T \{\Phi(\delta_\theta) - \Phi\}.\\ \nonumber
&\stackrel{\Delta}{=} I_1+I_2+I_3+I_4+I_5+I_6,\qquad\qquad\qquad\qquad\qquad~
\end{eqnarray}
on the set $\varepsilon_3 \cup \varepsilon_5 \cup \varepsilon_{13}$, we have
\begin{eqnarray}\label{thm2eq7}
||I_1||_\infty+||I_2||_\infty+||I_3||_\infty=O(\sqrt{n\log n}).
\end{eqnarray}

Define
\begin{eqnarray*}
\eta_1 = (\hat{Z} - Z)^T \{\mu-\Phi(\delta_\theta)\},
\eta_2 = (\hat{Z} - Z)^T (\hat{Z}_1-Z_1)\beta_1.
\end{eqnarray*}
Note that $\eta_{1M_{\beta}}=I_3$ in (\ref{term}), which we represent here using a second order Taylor expansion around $\alpha_1$,
\begin{eqnarray}\label{eta1r0}
I_3=X_{1\beta}^T W(\delta_\theta) \Delta X_{1\alpha} (\alpha_1 - \hat{\alpha}_1)+ \frac{1}{2}r_{I_3},
\end{eqnarray}
where $r_{I_3}$ in (\ref{eta1r0}) corresponds to second order remainder, whose $j$th component is given as
\begin{eqnarray*}
(\hat{\alpha}_1-\alpha_1)^T X_{1\alpha}^T W(\delta_\theta) \Sigma(\tilde{\alpha}) \diag(x^j) X_{1\alpha} (\hat{\alpha}_1-\alpha_1),
\end{eqnarray*}
where $\Sigma(\tilde{\alpha})$ is a diagonal matrix with the $i$th diagonal element $\pi^{''}(x_{1\alpha i}^T \tilde{\alpha})$ with $\tilde{\alpha}$ lying in the line segment between $\hat{\alpha}_1$ and $\alpha_1$. Since $\pi^{''}(\cdot)$ is a bounded function, we can bound $||r_{I_3}||_\infty$ by
\begin{eqnarray}\label{thm2eq3}
\max_j \displaystyle (\hat{\alpha}_1 - \alpha_1)^T X_{1\alpha}^T \diag(|W(\delta_\theta) x^j|) X_{1\alpha} (\hat{\alpha}_1 - \alpha_1),
\end{eqnarray}
whose order of magnitude is $O(s_\alpha n^{1-2\gamma_\alpha}\log^2 n)$ by (\ref{lamxmu-phi}).

We decompose $I_4$ in (\ref{term}) as $\eta_{2M_{\beta}}+Z_1^T (\hat{Z}_1-Z_1)\beta_1$. Using similar arguments, on the set $\varepsilon_9$, it follows from (\ref{lamxxbeta}) that
\begin{eqnarray}\nonumber
\displaystyle||Z_1^T (\hat{Z}_1-Z_1)\beta_1||_\infty &\le& \max_j(\hat{\alpha}_1 - \alpha_1)^T X_{1\alpha}^T \diag(|x^j\circ X\beta_0|) X_{1\alpha} (\hat{\alpha}_1 - \alpha_1)\\ \label{thm2eq4}
&+&||\xi_{4M_{\beta}}||_\infty=O(\sqrt{n\log n}+s_\alpha n^{1-2\gamma_\alpha}\log^2 n).
\end{eqnarray}

Using Taylor expansion, it is immediate to see that
\begin{eqnarray}\nonumber
\displaystyle||\eta_{2M_{\beta}}||_\infty &\le& \max_j(\hat{\alpha}_1 - \alpha_1)^T X_{1\alpha}^T \diag(|x^j\circ X\beta_0|) X_{1\alpha} (\hat{\alpha}_1 - \alpha_1)\\ \label{thm2eq5}
&=&O(s_\alpha n^{1-2\gamma_\alpha}\log^2 n),
\end{eqnarray}
by (\ref{lamxxbeta}). Combining (\ref{thm2eq4}) and (\ref{thm2eq5}) gives
\begin{eqnarray}\label{thm2eq1}
||\hat{Z}_1^T (\hat{Z}_1-Z_1)\beta_1||_\infty=O(\sqrt{n\log n})+O(s_\alpha n^{1-2\gamma_\alpha}\log^2 n).
\end{eqnarray}

So far, we have
\begin{eqnarray}\label{thm2eq6}
&||I_1+I_2+I_3+I_5+I_6-X_{1\beta}^T W(\delta_\theta) \Delta X_{1\alpha} (\alpha_1 - \hat{\alpha}_1)||_\infty\\ \nonumber
&=O(\sqrt{n\log n})+O(s_\alpha n^{1-2\gamma_\alpha}\log^2 n)+O(s_\beta n^{1-2\gamma_\beta}\log^2 n),
\end{eqnarray}
by (\ref{thm2eq7}), (\ref{eta1r0}), (\ref{thm2eq3}) and (\ref{thm2eq1}). Now we approximate $I_4$ by $X_{1\beta}^T \Delta X_{1\beta} (\delta_\beta-\beta_1)$ and bound the magnitude of error $||\omega_{M_{\beta}}||_\infty$ where $\omega=(\hat{Z}^T \hat{Z}_1 - X^T \Delta X_{1\beta}) (\delta_\beta - \beta_1)$. We present it as
\begin{eqnarray}\nonumber
	\omega_{M_{\beta}} &=& (\hat{Z}_1^T \hat{Z}_1 - X_{1\beta}^T \Delta X_{1\beta}) (\delta_\beta - \beta_1)=\hat{Z}_1^T (\hat{Z}_1-Z_1) (\delta_\beta - \beta_1)\\ \nonumber
	&+& (\hat{Z}_1-Z_1)^T Z_1 (\delta_\beta - \beta_1) +(Z_1^T Z_1 - X_{1\beta}^T \Delta X_{1\beta}) (\delta_\beta - \beta_1)\\ \label{omega}
	&\stackrel{\Delta}{=}& \omega_{1M_{\beta}} + \omega_{2M_{\beta}} + \xi_{5M_{\beta}} (\delta_\beta - \beta_1).
\end{eqnarray}

It follows from first-order Taylor expansion that the $j$th element in $\omega_{1M_{\beta}}$ can be presented as
\begin{eqnarray}\label{thm2eq8}
[(A-\hat{\pi}) \circ x^j \circ \{\Delta(\tilde{\alpha}_1) X_{1\alpha} (\hat{\alpha}_1-\alpha_1)\}]^T X_{1\beta} (\delta_\beta-\beta_1),
\end{eqnarray}
where $\Delta(\hat{\alpha}_1)$ is a diagonal matrix with the $i$th diagonal component $\pi(x_i, \tilde{\alpha}_1) (1-\pi(x_i, \tilde{\alpha}_1))$, where $\tilde{\alpha}_1$ lies between the line segment of $\hat{\alpha}_1$ and $\alpha_1$. We decompose $x^j$ as the Hadamard product of two vectors, denoted by $\bar{x}^j \circ \tilde{x}^j$, where $$\bar{x}^j=(\sqrt{\mbox{sgn}(x_1^j)|x_1^j|}, \dots, \sqrt{\mbox{sgn}(x_n^j)|x_n^j|}),$$ $$\tilde{x}^j=(\mbox{sgn}(x_1^j)\sqrt{\mbox{sgn}(x_1^j)|x_1^j|}, \dots, \mbox{sgn}(x_n^j)\sqrt{\mbox{sgn}(x_n^j)|x_n^j|}).$$

Let $\varphi=(A-\hat{\pi}) \circ \tilde{x}^j \circ \{\Delta(\tilde{\alpha}_1) X_{1\alpha} (\hat{\alpha}_1-\alpha_1)\}$, we have
\begin{eqnarray}
\label{thm2eq9}&&||[(A-\hat{\pi}) \circ \tilde{x}^j \circ \{\Delta(\tilde{\alpha}_1) X_{1\alpha} (\hat{\alpha}_1-\alpha_1)\}]^T X_{1\beta}||_2 ||\delta_\beta-\beta_1||_2\\
\nonumber&=& \sqrt{\varphi^T \diag(\bar{x}^j) X_{1\beta} X_{1\beta}^T \diag(\bar{x}^j) \varphi} ||\delta_\beta-\beta_1||_2\\
\nonumber&\le& \sqrt{\lambda_{\max}(X_{1\beta}^T \diag(|x^j|) X_{1\beta})} ||\delta_\beta-\beta_1||_2 ||\varphi||_2.
\end{eqnarray}

Since $||A-\hat{\pi}||_\infty \le 1$, elements in $\Delta(\tilde{\alpha}_1)$ are bounded, we have
\begin{eqnarray}\label{thm2eq10}
||\varphi||_2 &\le& ||\diag(\tilde{x}^j) X_{1\alpha}(\hat{\alpha}_1-\alpha_1)||_2 \\ \nonumber
&\le& \sqrt{\lambda_{\max}\{X_{1\alpha}^T \diag(|x^j|) X_{1\alpha}\}}||\hat{\alpha}_1-\alpha_1||_2.
\end{eqnarray}

Combining (\ref{thm2eq9}) with (\ref{thm2eq10}) gives
\begin{eqnarray}\label{thm2eq11}
	&\displaystyle||\omega_{1M_{\beta}}||_\infty \le \max_{j=1}^p \sqrt{\lambda_{\max}\{X_{1\alpha}^T \diag(|x^j|) X_{1\alpha}\} ||\hat{\alpha}_1 - \alpha_1||_2^2}\\ \nonumber
	&\displaystyle \max_{j=1}^p \sqrt{\lambda_{\max}\{X_{1\beta}^T \diag(|x^j|) X_{1\beta}\} ||\hat{\beta}_1-\beta_1||_2^2},
\end{eqnarray}
which is $O(\sqrt{s_\alpha s_\beta} n^{1-\gamma_\alpha-\gamma_\beta}\log^2 n)$ by (\ref{lamx1a}) and (\ref{lamx1b}).

By the same argument, we can verify that $||\omega_{2M_{\beta}}||_\infty$ is of the same order. Note that on the set $\varepsilon_{11}$,
\begin{eqnarray*}
||\xi_{5M_{\beta}} (\delta - \beta_1)||_\infty\le||\xi_{5M_{\beta}}||_\infty ||\delta - \beta_1||_\infty=O(s_\beta n^{1-2\gamma_\beta} \log^2 n),
\end{eqnarray*}
these together with (\ref{thm2eq11}), yields
\begin{eqnarray}\label{thm2eq2}
||\omega_{M_{\beta}}||_\infty=O(s_\alpha n^{1-2\gamma_\alpha}\log^2 n)+O(s_\beta n^{1-2\gamma_\beta}\log^2 n).
\end{eqnarray}

Define vector-valued function
\begin{eqnarray}
\nonumber\Psi_1(\delta_\beta, \delta_\theta)&=& B_{n\beta}^{-1} [\hat{Z}_1^T \{y - \Phi(\delta_\theta)- \hat{Z_1} \delta_\beta\} - n\lambda_{2n} \bar{\rho}_2(\delta_\beta)]\\
\nonumber&=& B_{n\beta}^{-1} \{I_1+I_2+I_3+I_4+I_5+I_6-n\lambda_{2n} \bar{\rho}_2(\delta_\beta)\}\\
\nonumber&=& \delta_\beta -\beta_1+B_{n\beta}^{-1} \{I_1+I_2+I_3+\omega_{M_{\beta}}+I_5+I_6-n\lambda_{2n} \bar{\rho}_2(\delta_\beta)\}\\
\label{thm2eq12}&\stackrel{\Delta}{=}& \delta_\beta-\beta_1+u_\beta,
\end{eqnarray}
then equation (\ref{betaeqcon}) is equivalent to $\Psi_1(\delta_\beta, \delta_\theta)=0$. It follows from (\ref{betapenalty}), (\ref{thm2eq6}) and (\ref{thm2eq2}) that
\begin{eqnarray*}
&\displaystyle||u_\beta||_\infty \le \sup_{\delta \in H_\theta}||B_{n\beta}^{-1} X_{1\beta}^T W(\delta) \Delta X_{1\alpha}(\hat{\alpha}_1-\alpha_1)||_\infty+||B_{n\beta}^{-1}||_\infty \\
\displaystyle&\{O(s_\alpha n^{1-2\gamma_\alpha}\log^2 n)+O(s_\beta n^{1-2\gamma_\beta}\log^2 n)+O(\sqrt{n\log n})+n\lambda_{2n}\rho^{'}(d_{n\beta})\}.
\end{eqnarray*}

By similar arguments in the proof of Theorem 2 in \cite{fan2011}, we have
\begin{eqnarray}\label{alphaeqcon}
&||B_{n\alpha}(\hat{\alpha}_1-\alpha_1)||_\infty=O(s_\alpha n^{1-2\gamma_\alpha}\log^2 n)+O(\sqrt{n\log n})\\ \nonumber
&+n\lambda_{1n}\rho_1^{'}(d_{n\alpha}),
\end{eqnarray}
on the set $\varepsilon_1 \cup \varepsilon_2$. Thus by (\ref{x1x1}), (\ref{xxa1norm}), (\ref{lambda}) and (\ref{lambda2}), we have
\begin{eqnarray*}
&||u_\beta||_\infty \le O[b_{\alpha\beta}\{s_\alpha n^{-2\gamma_\alpha} \log^2 n+\sqrt{\log n/n}+\lambda_{1n}\rho_1^{'}(d_{n\alpha})\}]\\
&+O[b_\beta\{s_\alpha n^{-2\gamma_\alpha} \log^2 n+s_\beta n^{-2\gamma_\beta} \log^2 n+\sqrt{\log n/n}+\lambda_{2n} \rho_2^{'}(d_{n\beta})\}].
\end{eqnarray*}

Therefore by (\ref{thm2eq11}), for sufficiently large $n$, if $(\delta_\beta - \beta_1)_j = n^{-\gamma_\beta} \log n$,
\begin{eqnarray}\label{thm2eq13}
	\Psi_{1j}(\delta_\beta, \delta_\theta) > 0,
\end{eqnarray}
and if $(\delta - \beta_1)_j = -n^{-\gamma_\beta} \log n$,
\begin{eqnarray}\label{thm2eq14}
	\Psi_{1j}(\delta_\beta, \delta_\theta) < 0.
\end{eqnarray}

Similarly we write the left-hand side of (\ref{thetaeqcon}) as
\begin{eqnarray}\label{term1}
(\hat{\phi}_1-\phi_1)^T (e-Z\beta_0) + \xi_{3M_{\theta}^\prime} + \hat{\phi}_1^T (\mu - \Phi) - \hat{\phi}_1^T (\hat{\Phi} - \Phi).
\end{eqnarray}

It is immediately to see that
\begin{eqnarray}\label{thm2eq15}
||\xi_{3M_{\theta}^\prime}||_\infty = O(\sqrt{n\log n}),
\end{eqnarray}
on the set $\varepsilon_{5}$. The $L_\infty$ norm of the first term in (\ref{term1}) is bounded by
\begin{eqnarray}\label{thm2eq16}
\sup_{\delta \in H_\theta}||\xi_{7M_{\beta}}(\delta)||_\infty=O(\sqrt{n\log n}),
\end{eqnarray}
on the set $\varepsilon_{15}$.


Using second-order Taylor expansion, we approximate the last term in (\ref{term1}) by its first-order term $\hat{\phi}_1^T \hat{\phi}_1 (\delta_\theta - \theta_1^\star)$. It follows from (\ref{lamphidphi}) that the $L_\infty$ norm of the remainder term is bounded from above by
\begin{eqnarray}\label{thm2eq17}
\displaystyle ~~~~\max_{l=1}^{s_\theta} \lambda_{\max} \left\{\frac{\partial (|\phi^l(\delta_\theta)|)^T \phi_1(\tilde{\delta}_\theta)}{\partial \theta_1}\right\}||\delta_\theta - \theta_1^\star||_2^2=O(s_\theta n^{1-2\gamma_\theta}\log^2 n),
\end{eqnarray}
where $\tilde{\delta}_\theta$ lies between the line segment of $\theta_1^\star$ and $\delta_\theta$.

Define $\Psi_2(\delta_\beta, \delta_\theta)=\{\phi_1(\delta_\theta)^T \phi_1(\delta_\theta)\}^{-1}[\phi_1(\delta_\theta)^T \{Y-\Phi(\delta_\theta)\}-n\lambda_{3n} \bar{\rho}_3(\delta_\theta)]$, equation (\ref{thetaeqcon}) is equivalent to $\Psi_2(\delta_\beta, \delta_\theta)=0$. Similarly to $\Psi_1(\delta_\beta, \delta_\theta)$, we now show $\Psi_2(\delta_\beta, \delta_\theta)$ is mainly dominated by $\delta_\theta-\theta_1^\star$. Define $u_\theta=\Psi_2(\delta_\beta, \delta_\theta)-\delta_\theta+\theta_1^\star$, it follows from (\ref{phimu1}), (\ref{phiphiinfty}), (\ref{btheta}), (\ref{term1}), (\ref{thm2eq15}), (\ref{thm2eq16}) and (\ref{thm2eq17}) that
\begin{eqnarray}\nonumber
||u_\theta||_\infty &\le& ||\Psi_2(\delta_\beta, \delta_\theta)-\delta_\theta+\theta_1^\star||_\infty\le ||\{\phi_1(\delta_\theta)^T \phi_1(\delta_\theta)\}^{-1}||_\infty\{||\xi_{3M_{\theta}^\prime}||_\infty\\
\nonumber&+& ||\xi_{7M_{\theta}^\prime}(\delta_\theta)||_\infty+||\Phi(\delta_\theta)-\Phi-\phi(\delta_\theta)^T(\delta_\theta-\theta_1^\star)||_\infty+ n\lambda_{3n} \rho_3^{'}(d_{n\theta})\}\\
\nonumber&+& ||\{\phi_1(\delta_\theta)^T \phi_1(\delta_\theta)\}^{-1}\phi_1(\delta_\theta)^T (\mu-\Phi)||_\infty\\
\label{thm2eq30}&=& o(n^{-\gamma_\theta}\log n)+O(n^{-\gamma_\theta}\log n).
\end{eqnarray}

Therefore, we can find a large constant $K < \infty$, for $n$ large enough such that if $(\delta_\theta - \theta_1^\star)_j = Kn^{-\gamma_\theta} \log n$,
\begin{eqnarray}\label{thm2eq18}
	\Psi_2(\delta_\beta, \delta_\theta) > 0,
\end{eqnarray}
and if $(\delta_\theta - \theta_1^\star)_j = -Kn^{-\gamma_\theta} \log n$,
\begin{eqnarray}\label{thm2eq19}
	\Psi_2(\delta_\beta, \delta_\theta) < 0.
\end{eqnarray}

Combining (\ref{thm2eq13}), (\ref{thm2eq14}) with (\ref{thm2eq18}) and (\ref{thm2eq19}), an application of Miranda's existence theorem shows equations (\ref{betaeqcon}), (\ref{thetaeqcon}) have a solution $(\hat{\beta}_1, \hat{\theta}_1)$ in $\aleph$.
\quad\newline

\textit{Step 2}. Let $(\hat{\beta}^T, \hat{\theta}^T)^T \in \aleph$ a solution to equations (\ref{betaeqcon}) and (\ref{thetaeqcon}) with $\hat{\beta}_{M_{\beta}^c} = 0$ and $\hat{\theta}_{M_{\beta}^c}=0$. We show that $(\hat{\beta}^T, \hat{\theta}^T)^T$ satisfies inequalities (\ref{betaineqcon}) and (\ref{thetaineqcon}). Decompose (\ref{betaineqcon}) as the sum of the following terms,
\begin{eqnarray}\nonumber
&\hat{Z}_2^T (Y - \hat{\Phi} - \hat{Z}^T_1 \hat{\beta}_1) = \xi_{1M_{\beta}^c} + \xi_{2M_{\beta}^c} + Z_2^T(\hat{Z}_1 - Z_1)\beta_1 +\xi_{5M_{\beta}^c} (\hat{\beta}_1 - \beta_1)+\omega_{1M_{\beta}^c} \\ \label{thm2eq20}
& + \omega_{2M_{\beta}^c} + \eta_{1M_{\beta}^c} + X_{2\beta}^T \Delta X_{1\beta} (\hat{\beta}_1 - \beta_1) + \eta_{2M_{\beta}^c} - Z_2 (\hat{\Phi} - \Phi),
\end{eqnarray}
on the set $\varepsilon_4\cup\varepsilon_6\cup\varepsilon_{10}\cup\varepsilon_{12}$, it is immediately to see that
\begin{eqnarray}\label{thm2eq21}
&||\xi_{1M_{\beta}^c}||_\infty+||\xi_{2M_{\beta}^c}||_\infty+||\xi_{5M_{\beta}^c}(\hat{\beta}_1 - \beta_1)||_\infty+||Z_2^T (\hat{\Phi}-\Phi)||_\infty\\ \nonumber
&=O(n^{1-d_\beta}\sqrt{\log n}).
\end{eqnarray}

By (\ref{lamx1a}), (\ref{lamx1b}) and (\ref{thm2eq11}), a first-order Taylor expansion gives
\begin{eqnarray}\label{thm2eq22}
~~~~~~~~~~~||\omega_{1M_{\beta}^c}||_\infty+||\omega_{2M_{\beta}^c}||_\infty=O(s_\alpha n^{1-2\gamma_\alpha}\log^2 n)+O(s_\beta n^{1-2\gamma_\beta}\log^2 n).
\end{eqnarray}

Similarly, it follows from (\ref{lamxxbeta}) and (\ref{thm2eq5}) that
\begin{eqnarray}\label{thm2eq23}
||\eta_{2M_{\beta}}||_\infty = O(s_\alpha n^{1-2\gamma_\alpha} \log^2 n).
\end{eqnarray}

On the set $\varepsilon_{10}$, by (\ref{lamxxbeta}) and (\ref{thm2eq4}), we have
\begin{eqnarray}\label{thm2eq24}
~~~~~~||Z_2^T (\hat{Z}_1-Z_1)\beta||_\infty=O(n^{1-d_\beta}\sqrt{\log n})+O(s_\alpha n^{1-2\gamma_\alpha}\log^2 n).
\end{eqnarray}

Approximating $\eta_{1M_{\beta}^c}$ by $X_{2\beta}^T W(\delta_\theta) \Delta X_{1\alpha} (\alpha_1 - \hat{\alpha}_1)$, the $L_\infty$ norm of remainder error term is bounded from above by
\begin{eqnarray}\label{thm2eq25}
~~~~~~~~~~~~~~\displaystyle (\hat{\alpha}_1 - \alpha_1)^T X_{1\alpha}^T \diag(|W(\delta_\theta) x^j|) X_{1\alpha} (\hat{\alpha}_1 - \alpha_1)=O(s_\alpha n^{1-\gamma_\theta}\log^2 n),
\end{eqnarray}
by (\ref{lamxmu-phi}).
Let $u_\beta^\prime=\hat{Z}_2^T (Y - \hat{\Phi} - \hat{Z}^T_1 \hat{\beta}_1)-X_{2\beta}^T \Delta X_{1\beta} (\hat{\beta}_1 - \beta_1)-X_{2\beta}^T W(\hat{\theta}_1) \Delta X_{1\alpha} (\alpha_1 - \hat{\alpha}_1)$, it follows from (\ref{thm2eq20})--(\ref{thm2eq25}) that
\begin{eqnarray}\label{thm2eq27}
~~~~~~||u_\beta^\prime||_\infty=O(n^{1-d_\beta}\sqrt{\log n}+s_\alpha n^{1-2\gamma_\alpha}\log^2 n+s_\beta n^{1-2\gamma_\beta}\log^2 n).
\end{eqnarray}

Since $\hat{\beta}_1$ solves (\ref{betaeqcon}), we have
\begin{eqnarray}\label{thm2eq26}
\hat{\beta}_1-\beta_1=-u_\beta,
\end{eqnarray}
where $u_\beta$ is defined as $\Psi_1(\hat{\beta}_1, \hat{\theta}_1)+\beta_1-\hat{\beta}_1$. Combining (\ref{thm2eq26}) with (\ref{alphaeqcon}) and (\ref{thm2eq27}) gives
\begin{eqnarray}
\nonumber
&&\displaystyle ||\frac{1}{n\lambda_{2n}} \hat{Z}_2^T (Y -\hat{\Phi}- \hat{Z}^T_1 \hat{\beta}_1)||_\infty \le \frac{1}{n\lambda_{2n}}[||u_\beta^\prime||_\infty+||X_{2\beta}^T \Delta X_{1\beta} (X_{1\beta}^T \Delta X_{1\beta})^{-1}||_\infty\\ \nonumber
&&\{u_\beta-X_{1\beta}^T W(\hat{\theta}_1) \Delta X_{1\alpha} (\alpha_1 - \hat{\alpha}_1)\}+||X_{2\beta}^T W_\beta W(\hat{\theta}_1) X_{1\alpha}(X_{1\alpha}^T \Delta X_{1\alpha})^{-1}||_\infty\\ \nonumber
&&\{n\lambda_{1n} \rho_1^{'}(d_{n\alpha})+O(\sqrt{n\log n})+O(s_\alpha n^{1-2\gamma_\alpha}\log^2 n)\}] \le o(1)+C\rho_2^{'}(0+),
\end{eqnarray}
by (\ref{x2bx1b}), (\ref{x2bx1}), (\ref{lamn}) and (\ref{dbeta0}). Since $C<1$, for sufficiently large $n$, (\ref{betaineqcon}) is satisfied.

Now we verify (\ref{thetaineqcon}), decomposing $\hat{\phi}_2^T(Y-\hat{\Phi})$ as the sums of
\begin{eqnarray}\label{thm2eq28}
(\hat{\phi}_2 - \phi_2)^T (e-Z\beta_0) +\xi_{3M_{\theta}^{\prime c}}+\hat{\phi}_2^T (\mu-\Phi)+\hat{\phi}_2^T (\Phi-\hat{\Phi}),
\end{eqnarray}
on the set $\varepsilon_8 \cup \varepsilon_{16}$, we have
\begin{eqnarray}\label{thm2eq29}
||\xi_{3M_{\theta}^{\prime c}}||_\infty+||(\hat{\phi}_2 - \phi_2)^T (e-Z\beta_0)||_\infty = O(n^{1-d_\theta}\sqrt{\log n}).
\end{eqnarray}

Similar to (\ref{thm2eq17}), a second-order Taylor expansion gives
\begin{eqnarray}
||\hat{\phi}_2^T (\hat{\Phi} - \Phi) - \hat{\phi}_2^T \hat{\phi}_1(\hat{\theta}_1 - \theta_1^\star)||_\infty = O(s_\theta n^{1-2\gamma_\theta}\log^2 n),
\end{eqnarray}
by (\ref{lamphidphi}). Since $(\hat{\beta}_1, \hat{\theta}_1)$ is the solution to $\Psi_2(\delta_\beta, \delta_\theta)=0$, it follows from (\ref{thm2eq30}) that
\begin{eqnarray}\label{thm2eq31}
&||\hat{\phi}_2^T \hat{\phi}_1(\hat{\theta}_1-\theta_1^\star)-\hat{\phi}_2^T \hat{\phi}_1 (\hat{\phi}_1^T \hat{\phi}_1)^{-1}(\mu-\Phi)||_\infty\\ \nonumber
&= ||\hat{\phi}_2^T \hat{\phi}_1 (\hat{\phi}_1^T \hat{\phi}_1)^{-1}||_\infty \{O(\sqrt{n\log n}+s_\theta n^{1-2\gamma_\theta}\log^2 n)+n\lambda_{3n}\rho^{'}_3(d_{n\theta})\}.
\end{eqnarray}

By (\ref{thm2eq28})--(\ref{thm2eq31}) and conditions in (\ref{phimu2}), (\ref{phi2phi1infty}), (\ref{lambda2}) and (\ref{dtheta0}), the left-hand side of (\ref{thetaineqcon}) can be bounded by
\begin{eqnarray*}
	&&\frac{1}{n\lambda_{3n}}\{O(n^{1-d_\theta}\sqrt{\log n})+O(s_\theta n^{1-2\gamma_\theta}\log^2 n)\}+\frac{1}{n\lambda_{3n}}||\hat{\phi}_2^T \hat{\phi}_1 (\hat{\phi}_1^T \hat{\phi}_1)^{-1}||_\infty\\
	&&\{O(\sqrt{n\log n})+O(s_\theta n^{1-2\gamma_\theta}\log^2 n)+n\lambda_{3n}\rho^{'}_3(d_{n\theta})\}+\frac{1}{n\lambda_{3n}}\\
	&&||\hat{\phi}_2^T \{I-P_{\phi_1}(\hat{\theta}_1)\}(\mu-\Phi)||_\infty=o(1) + C\rho_{3}^{'}(0+),
\end{eqnarray*}
for $C<1$. Therefore (\ref{thetaineqcon}) is satisfied.
\quad\newline

\textit{Step 3.} Now we show the second order conditions (\ref{betasdcon}) and (\ref{thetasdcon}) hold. Because (\ref{thetasdcon}) is directly implied by (\ref{lam3}), it suffices to show that $\lambda_{\min}(\hat{Z}_1^T \hat{Z}_1) \ge \lambda_{\min}(X_{1\beta}^T \Delta X_{1\beta})$ for sufficiently large $n$. Since $(\hat{Z}_1 - Z_1)^T (\hat{Z}_1 - Z_1)$ is positive semi-definite, we have
\begin{eqnarray}\label{thm2eq32}
&\lambda_{\min}(\hat{Z}_1^T \hat{Z}_1) \ge \lambda_{\min}(X_{1\beta}^T \Delta X_{1\beta})\\ \nonumber
&+ \lambda_{\min}\{(\hat{Z}_1 - Z_1)^T Z_1 + Z_1^T (\hat{Z}_1 - Z_1)+\xi_{5M_{\beta}}\}.
\end{eqnarray}

Since any symmetric matrix $\Psi$, the absolute value of minimum eigenvalue can be bounded by $$|\lambda_{\min}(\Psi)| \le \sqrt{\lambda_{\max}(\Psi^2)} \le \sqrt{||\Psi||_\infty||\Psi||_1}=||\Psi||_\infty,$$
(\ref{betasdcon}) follows if we can show $||\xi_{5M_{\beta}}+(\hat{Z}_1 - Z_1)^T Z_1 + Z_1^T (\hat{Z}_1 - Z_1)||_\infty=o(n)$. But this is immediate to see because
$$||\xi_{5M_{\beta}}||_\infty=O(n^{1/2+\gamma_\beta}/\sqrt{\log n})=o(n),$$ on the set $\varepsilon_{11}$. Similar to (\ref{thm2eq11}), $||(\hat{Z}_1 - Z_1)^T Z_1 + Z_1^T (\hat{Z}_1 - Z_1)||_\infty$ can be bounded from above by
\begin{eqnarray}\label{thm2eq33}\\ \nonumber
\displaystyle 2\max_{j}\sqrt{s_\beta\lambda_{\max}\{X_{1\beta}^T \diag(|x^j) X_{1\beta}\} \lambda_{\max}\{X_{1\alpha}^T \diag(|x^j|) X_{1\alpha}\}||\hat{\alpha}_1-\alpha_1||_2^2},
\end{eqnarray}
which is $O(\sqrt{s_\alpha s_\beta}n^{1-\gamma_\alpha}\log n)=o(n)$ implied by the constrain $\max(l_1, l_2)< \gamma_\alpha$. This completes the proof.
\newpage
\section{Proof of Theorem 2}

We modify the right-hand side orders in $\varepsilon_{13}$ as $O(\sqrt{n})$, and redefine
\begin{eqnarray*}
\displaystyle \varepsilon_{10}=\max_{j=1}^{s_\beta}||\xi_{4j}||_\infty=O(\sqrt{n\log n}),\varepsilon_{11}=\max_{j=s_\beta}^p||\xi_{4j}||_\infty=O(n^{1-d_\beta}\sqrt{\log n}),\\
\displaystyle \varepsilon_{12}=\max_{j=1}^{s_\beta}||\xi_{5j}||_\infty=O(\sqrt{n\log n}),\varepsilon_{13}=\max_{j=s_\beta}^p||\xi_{5j}||_\infty=O(n^{1-d_\beta}\sqrt{\log n}).
\end{eqnarray*}

We still have $\prob(\varepsilon|X)\to 1$ due to the $L_2$ constraints in Condition 4.1. Convergence rate of $\hat{\alpha}_1$ follows directly from Theorem 3 in \cite{fan2011}, it suffices to show the convergence rate of $\hat{\beta}$ under $\varepsilon$. The proof is divided into two steps. In the first step, we establish the $\sqrt{s_\beta/n}$ consistency of $\hat{\beta}_1$ in the $s_\beta$-dimensional subspaces. In the second step, we show that the estimator is indeed a local minimizer, which satisfies (\ref{betaineqcon}) and (\ref{betasdcon}).

\textit{Step 1}. We constrain $(\ref{beta})$ on the subspace $\left\{\beta:\beta \in \mathbb{R}^p, \beta_{M_{\beta}^c}=0\right\}$, yielding the following constrained loss function
\begin{eqnarray*}
\displaystyle \bar{Q}_n(\delta_\beta) = \frac{1}{2n} \sum_{i=1}^n (Y_i - \hat{\Phi} - \delta_\beta^T x_{iM_{\beta}} (A_i - \hat{\pi}_i))^2
 + \sum_{j=1}^{s_\beta} \lambda_{2n} \rho_2(|\delta_{\beta j}|),
\end{eqnarray*}
where $\delta_\beta = (\delta_{\beta1}, \dots, \delta_{\beta s_\beta})^T$. We now show that there exists a strict local minimizer $(\hat{\beta}_1, \hat{\theta}_1)$ of $\bar{Q}_n(\delta_\beta)$ such that $||\hat{\beta}_1 - \beta_1||_2 = O_p(\sqrt{s_\beta/n})$. Consider the event
\begin{eqnarray*}
H_n = \left\{\bar{Q}_n(\beta_1) < \min_{\delta \in \partial N_\tau} \bar{Q}_n(\delta_\beta)\right\},
\end{eqnarray*}
where $\partial N_\tau$ denotes the boundary of the closed set $$N_\tau = \left\{\delta_\beta: ||\delta_\beta - \beta_1||_2 \le \sqrt{s_\beta/n}\tau\right\}$$ with $\tau \in (0, \infty)$. Clearly, on the event $H_n$, there exists a local minimizer $\hat{\beta}_1$ of $\bar{Q}_n(\delta_\beta)$ in $N_\tau$. Thus, we only need to show $\prob(H_n|\varepsilon) \to 1$ as $n \to \infty$.

Let $n$ be sufficiently large such that $\sqrt{s_\beta/n}\tau \le d_{n\beta}$ since $d_{n\beta} \gg \sqrt{s_\beta/n}$ by Condition $5^\prime$. It is easy to see that $\delta_\beta \in N_\tau$ entails sgn$(\delta_\beta)$ = sgn$(\beta_1)$, $||\delta_\beta - \beta_1||_\infty \le d_{n\beta}$ and $\min_j |\delta_{\beta j}| \ge d_{n\beta}$.

Rewrite $\bar{Q}_n(\beta_1) - \bar{Q}_n(\delta_\beta)$ using second-order Taylor's expansion, we have
\begin{eqnarray}\label{term3}
&(\delta_\beta - \beta_1)^T \{\frac{1}{n} \hat{Z}_1^T (Y - \hat{\Phi} - \hat{Z}_1 \beta_1) - \lambda_{2n} \bar{\rho}_2 (\beta_1)\}\\ \nonumber
&-\frac{1}{2}(\delta_\beta - \beta_1)^T(\hat{Z}_1^T \hat{Z}_1/n  + \Lambda_\beta) (\delta_\beta - \beta_1),
\end{eqnarray}
where $\Lambda_\beta$ is a diagonal matrix with maximum absolute element bounded by $\lambda_{2n} \kappa_\beta$.

We first show the minimum eigenvalue of $\hat{Z}_1^T \hat{Z}_1/n  + \Lambda_\beta$ is bounded from below by a positive number. By (\ref{thm2eq32}) and conditions in (\ref{lamminx}) and (\ref{kappao1}), it suffices to show
\begin{eqnarray}\label{term4}
||(\hat{Z}_1 - Z_1)^T Z_1+Z_1^T (\hat{Z}_1 - Z_1)||_\infty + ||\xi_{5M_{\beta}}||_\infty = o(n).
\end{eqnarray}
By (\ref{thm2eq33}) and conditions in (\ref{lamx1a}) and (\ref{lamx1b}), $||(\hat{Z}_1-Z_1)^T Z_1||_\infty+||Z_1^T (\hat{Z}_1 - Z_1)||_\infty=\sqrt{s_\alpha s_\beta n}$, which is $o(n)$. $||\xi_{5M_{\beta}}||_\infty = o(n)$ follows under $\varepsilon_{11}$. (\ref{term4}) together with (\ref{lamminx}) entails, for sufficiently large $n$, that
\begin{eqnarray*}
\lambda_{\min}(\frac{1}{n} \hat{Z}_1^T \hat{Z}_1 + \Lambda_\beta) \ge b_2 - \frac{b_2}{2} = \frac{b_2}{2}.
\end{eqnarray*}

Therefore,
\begin{eqnarray}\label{Q1}
&\bar{Q}_n(\beta_1) - \bar{Q}_n(\delta_\beta) \le (\delta_\beta-\beta_1)^T [v_\beta-\frac{1}{4n}\{B_{n\beta}+o(1)\} (\delta_\beta-\beta_1)]\\ \nonumber
&+\frac{1}{n}(\delta_\beta-\beta_1)^T X_{1\beta}^T W(\hat{\theta}_1)\Delta X_{1\alpha} (\alpha_1-\hat{\alpha}_1)-\frac{1}{4n}(\delta_\beta-\beta_1)^T B_{n\beta}(\delta_\beta-\beta_1),
\end{eqnarray}
where $\displaystyle v_\beta = \frac{1}{n} \{\hat{Z}_1^T (Y - \hat{\Phi} - \hat{Z} \beta_0) + X_{1\beta}^T \diag(\mu-\hat{\Phi}) \Delta X_{1\alpha} (\hat{\alpha}_1-\alpha_1)\} - \lambda_{2n} \bar{\rho}_2 (\beta_1)$.

We next show the second line in (\ref{Q1}) is smaller than $0$ for sufficiently large $\tau$. It follows from Cauchy-Schwartz inequality that
\begin{eqnarray*}
&(\delta_\beta-\beta_1)^T X_{1\beta}^T \diag(\mu-\hat{\Phi}) \Delta X_{1\alpha} (\alpha_1-\hat{\alpha}_1)\le ||B_{n\beta}^{1/2} (\delta_\beta-\beta_1)||_2\\ &||B_{n\beta}^{-1/2} X_{1\beta}^T \diag(\mu-\hat{\Phi}) \Delta X_{1\alpha} (\hat{\alpha}_1-\alpha_1)||_2,
\end{eqnarray*}
by which we only need to show
\begin{eqnarray}\label{Q2}
~~||B_{n\beta}^{1/2} (\delta_\beta-\beta_1)||_2 \ge ||B_{n\beta}^{-1/2} X_{1\beta}^T \diag(\mu-\hat{\Phi}) \Delta X_{1\alpha} (\hat{\alpha}_1-\alpha_1)||_2.
\end{eqnarray}
It follows by (\ref{lamminx}) that the left-hand side of (\ref{Q2}) is larger than $O(\tau \sqrt{s_\beta})$. Therefore for sufficiently large $\tau$, it suffices to show its right-hand side is $O(\sqrt{s_\beta})$. We present $\hat{\alpha}_1-\alpha_1$ by
\begin{eqnarray}\label{alpha1hat-alpha1}
B_{n1}^{-1}\{X_{1\alpha}^T (A-\pi)+n\lambda_{1n}\bar{\rho}_1(\hat{\alpha}_1)+r\},
\end{eqnarray}
with $r$ whose $L_\infty$ norm is $O(s_\alpha)$.

Since
\begin{eqnarray*}
&&\Mean ||B_{n\alpha}^{-1/2} X_{1\alpha} (A-\pi)||_2^2 \\
&=& \Mean \{\mbox{tr}(B_{n\alpha}^{-1/2} X_{1\alpha} (A-\pi) (A-\pi)^T X_{1\alpha}^T B_{n\alpha}^{-1/2})\}\\
&=& \mbox{tr} (E_{s_\alpha}) = s_\alpha,
\end{eqnarray*}
we have
\begin{eqnarray}\label{thm3eq13}
||B_{n1}^{-1/2} X_{1\alpha} (A-\pi)||_2=O_p(\sqrt{s_\alpha}).
\end{eqnarray}

It follows from (\ref{lamminx}) and (\ref{lambdadn}) that
\begin{eqnarray}\label{thm3eq14}
&&||B_{n1}^{-1/2} \{n\lambda_{1n}\bar{\rho}_1(\hat{\alpha}_1)+r\}||_2 \le \frac{1}{\sqrt{n}} ||n\lambda_{1n}\bar{\rho}_1(\hat{\alpha}_1)+r||_2\\ \nonumber
&\le& \frac{2}{\sqrt{n}} \{||\lambda_{1n}\bar{\rho}_1(d_{n\alpha})||_2 + ||r||_2\}=O(\sqrt{s_\alpha}).
\end{eqnarray}

Combining (\ref{thm3eq13}) together with (\ref{thm3eq14}) we have
\begin{eqnarray*}
||B_{n\beta}^{-1/2}(\hat{\alpha}_1-\alpha_1)||_2=O_p(\sqrt{s_\alpha}),
\end{eqnarray*}
which along with condition in (\ref{xx1a2norm}) gives
\begin{eqnarray*}
||B_{n\beta}^{-1/2} X_{1\beta}^T \diag(\mu-\hat{\Phi}) \Delta X_{1\alpha} (\hat{\alpha}_1-\alpha_1)||_2=O(\sqrt{s_\beta}),
\end{eqnarray*}
thus (\ref{Q2}) is satisfied.

The next step is to prove that the quantity in the first line of (\ref{Q1}) is less than $0$, an application of Markov's inequality entails
\begin{eqnarray*}
\prob(H_{n}|\varepsilon) \ge \prob(||v_\beta||_2^2 < \frac{s_\beta b_2^2 \tau^2}{64n}|\varepsilon) \ge 1 - \frac{64n\Mean ||v_\beta||_2^2 I_{\varepsilon}}{s_\beta b_2^2 \tau^2},
\end{eqnarray*}
by which it suffices to show that $\Mean||v_\beta||_2^2I_\varepsilon = O(\frac{s_\beta}{n})$.

We rewrite $nv_\beta$ as the sums of the following terms,
\begin{eqnarray}\nonumber
&nv_\beta = \xi_{1M_{\beta}} + \xi_{2M_{\beta}} + Z_1^T (\hat{Z}_1 - Z_1) \beta_1 + \eta_{2\mathbb{R}_{0}} +\eta_{3M_{\beta}} - Z_1^T\{\Phi(\delta_\theta) - \Phi\}\\ \label{thm3eq15}
&-n\lambda_{2n} \bar{\rho}_2(\beta_1)+\eta_{1M_{\beta}}+X_{1\beta}^T W(\hat{\theta}_1) \Delta X_{1\alpha} (\hat{\alpha}_1-\alpha_1).
\end{eqnarray}

It is immediate to see that
\begin{eqnarray}\label{thm3eq2}
\Mean ||\xi_{1M_{\beta}}||_2^2I_\varepsilon \le \Mean ||\xi_{1M_{\beta}}||_2^2 = \sigma^2 \mbox{tr}(\hat{Z}_1^T \hat{Z}_1) = O(s_\beta n),
\end{eqnarray}
since $\mbox{tr}(\hat{Z}_1^T \hat{Z}_1) \le \mbox{tr}(X_{1\beta}^T X_{1\beta})=s_\beta n$ considering that the covariates are standardized. On the set $\varepsilon_{15}$,
\begin{eqnarray}\label{thm3eq3}
\displaystyle\Mean||Z_1^T(\hat{\Phi} - \Phi)||_2^2I_\varepsilon\le s_\beta ||Z_1^T(\hat{\Phi} - \Phi)||_\infty^2=O(s_\beta n).
\end{eqnarray}

It follows from (\ref{tracexmu}) that
\begin{eqnarray}\label{thm3eq4}
\Mean ||\xi_{2M_{\beta}}||_2^2I_\varepsilon \le \Mean ||\xi_{2M_{\beta}}||_2^2=\mbox{tr}(X_{1\beta}^T W\Delta W X_{1\beta})=O(s_\beta n).
\end{eqnarray}

A first-order Taylor expansion gives
\begin{eqnarray}\label{thm3eq5}
&\displaystyle\Mean ||\eta_{2M_{\beta}}||_2^2I_\varepsilon \le s_\beta ||\eta_{2M_{\beta}}||_\infty^2I_\varepsilon \\ \nonumber
&\le s_\beta [\displaystyle \max_j \lambda_{\max}\{X_{1\alpha}^T \diag(|x^j\circ X\beta_0|)X_{1\alpha}\} ||\hat{\alpha}_1 - \alpha_1||_2^2]^2I_\varepsilon=O(s_\beta s_\alpha^2),
\end{eqnarray}
by (\ref{lamxxbeta}). Expand $Z_1^T (\hat{Z}_1-Z_1)\beta_1$ to the second order using Taylor expansion around $\alpha_1$ gives
\begin{eqnarray}\label{thm3eq6}
&\displaystyle ||Z_1^T (\hat{Z}_1 - Z_1) \beta_1||_2^2I_\varepsilon\le 2||\xi_{3M_{\beta}}||_2^2 ||\hat{\alpha}_1 - \alpha_1||_2^2I_\varepsilon+\\ \nonumber
&\displaystyle 2s_\beta [\max_{j=1}^p \lambda_{\max}\{X_{1\alpha}^T \diag(|x^j\circ X \beta_0|)X_{1\alpha}\} ||\hat{\alpha}_1 - \alpha_1||_2^2]^2I_\varepsilon=O(s_\beta s_\alpha^2 \log n),
\end{eqnarray}
by (\ref{lamxxbeta}). Similarly, it follows from (\ref{lamxmu-phi}) that
\begin{eqnarray}\label{thm3eq7}
&||\eta_{1M_{\beta}}+X_{1\beta}^T W(\hat{\theta}_1) \Delta X_{1\alpha} (\hat{\alpha}_1-\alpha_1)||_2^2I_\varepsilon \le s_\beta\\ \nonumber
&\displaystyle  [\max_j \lambda_{\max}\{X_{1\alpha}^T \diag(|W(\hat{\theta}_1)x^j|)  X_{1\alpha}\} ||\hat{\alpha}_1 - \alpha_1||_2^2]^2=O(s_\alpha^2 s_\beta).
\end{eqnarray}

By the monotonicity of $\rho^\prime_2(\cdot)$ and the condition in (\ref{lambdadn}), we have
$\displaystyle ||\lambda_{2n}\rho_2^\prime(\hat{\beta}_1)||_2^2 \le ||\lambda_{2n}\rho_2^\prime(d_{n\beta})||_2^2 = O(s_\beta n)$. This along with (\ref{thm3eq13})--(\ref{thm3eq7}), and the condition on the nonsparsity size $\max(l_1,l_2)<\frac{1}{2}$, gives $\Mean||v_\beta||_2^2I_\varepsilon = O(\frac{s_\beta}{n})$. Therefore the convergence rate of $\hat{\beta}_1$ is established.
\quad\newline

\textit{Step 2}. It remains to show that the vector $\hat{\beta}=(\hat{\beta}_1^T, 0)^T$ is indeed a strict local minimizer of $\bar{Q}_n(\delta_\beta)$ on the space $\mathbb{R}^p \times \mathbb{R}^q$, which need to check condition (\ref{betaineqcon}) and (\ref{betasdcon}).

On the set $\varepsilon_2 \cup \varepsilon_4 \cup \varepsilon_{16}$, we have
\begin{eqnarray}\label{thm3eq8}
&&\displaystyle||\xi_{1M_{\beta}^c}||_\infty = ||\xi_{2M_{\beta}^c}||_\infty =||Z_2^T (\hat{\Phi} - \Phi)||_\infty = O(n^{1-d_\beta}\sqrt{\log n}).
\end{eqnarray}

Besides, on $\varepsilon_{12}$, we have
\begin{eqnarray}&\nonumber
\displaystyle||\xi_{5M_{\beta}^c} (\hat{\beta}_1 - \beta_1)||_\infty\le \max_{j=s_\beta+1}^p |\xi_{5j}^T (\hat{\beta}_1-\beta_1)|\le \max_{j=s_\beta+1}^p ||\xi_{5j}||_2 ||\hat{\beta}_1-\beta_1||_2\\ \label{thm3eq16} & \displaystyle \le \max_{j=s_\beta+1}^p \sqrt{s_\beta} ||\xi_{5j}||_\infty ||\hat{\beta}_1-\beta_1||_2=O(n^{1-d_\beta}\sqrt{\log n}),
\end{eqnarray}
since $s_\beta\ll \sqrt{n}$. It follows from (\ref{lamx1a}) and (\ref{lamx1b}) that
\begin{eqnarray}\nonumber
&||\omega_{1M_{\beta}^c}||_\infty\le \displaystyle \max_{j=1}^p \sqrt{s_\beta \lambda_{\max}\{X_{1\alpha}^T\diag(|x^j|)X_{1\alpha}\}\{X_{1\beta}^T\diag(|x^j|)X_{1\beta}\}}\\ \label{thm3eq9}
&||\hat{\alpha}_1 - \alpha_1||_2 ||\hat{\beta}_1 - \beta_1||_2 = O(\sqrt{s_\alpha} s_\beta)=O(\sqrt{s_\beta n}).
\end{eqnarray}

With a similar argument,
\begin{eqnarray}\label{thm3eq10}
||\omega_{2M_{\beta}^c}||_\infty=O(\sqrt{s_\beta n}).
\end{eqnarray}

On the set $\varepsilon_{10}$, it follows from (\ref{lamxxbeta}) that
\begin{eqnarray}\nonumber
&||Z_2^T (\hat{Z}_1 - Z_1)\beta_1||_\infty \le \displaystyle \max_{j=1}^p \lambda_{\max}\{X_{1\alpha}^T \diag(|x^j\circ X \beta_0|)X_{1\alpha}\} ||\hat{\alpha}_1 - \alpha_1||_2^2\\ &\label{thm3eq11}
+||\xi_{4M_{\beta}^c} (\hat{\alpha}_1 - \alpha_1)||_\infty= o(\sqrt{n}) + o(n^{1-d_\beta}\sqrt{\log n}).
\end{eqnarray}

Similarly, we have
\begin{eqnarray}\label{thm3eq12}
\qquad||\eta_{2M_{\beta}^c}||_\infty \le \displaystyle \lambda_{\max}\{X_{1\alpha}^T \diag|x^j \circ X\beta_0|X_{1\alpha}\}||\hat{\alpha}_1 - \alpha_1||_2^2=o(\sqrt{n}),
\end{eqnarray}

By (\ref{lamxmu-phi}), a second-order Taylor expansion gives
\begin{eqnarray}\label{thm3eq17}
&||\eta_{1M_{\beta}^c}-X_{2\beta}^T W(\hat{\theta}_1) \Delta X_{1\alpha} (\hat{\alpha}_1-\alpha_1)||_\infty \\ \nonumber
&\le \displaystyle\max_{j} \lambda_{\max}\{X_{1\alpha}^T \diag(|W(\hat{\theta}_1) x^j|)X_{1\alpha}\}||\hat{\alpha}_1 - \alpha_1||_2^2 = o(\sqrt{n}).
\end{eqnarray}

It follows from (\ref{thm2eq26}) and (\ref{thm3eq11}) that
\begin{eqnarray}\label{thm3eq18}
\hat{\beta}_1-\beta_1=-B_{n\beta}^{-1}\{nv_\beta-X_{1\beta}^T W(\hat{\theta}_1) \Delta X_{1\alpha} (\hat{\alpha}_1-\alpha_1)+\omega_{M_{\beta}}\}.
\end{eqnarray}

On the set $\varepsilon_3\cup\varepsilon_5$, it follows from (\ref{thm2eq11}) and $\max(l_1, l_2)<\frac{1}{2}$ that
\begin{eqnarray}
	\label{thm4eq7}&&||\omega_{1M_{\beta}}||_2 \le \sqrt{s_\beta} ||\omega_{1M_{\beta}}||_\infty = O_p(\sqrt{s_\alpha}s_\beta) = o_p(\sqrt{s_\beta n}),\\
	\label{thm4eq9}&&||\omega_{2M_{\beta}}||_2\le \sqrt{s_\beta}||\omega_{2M_{\beta}}||_\infty=O_p(\sqrt{s_\alpha}s_\beta) = o_p(\sqrt{s_\beta n}).
\end{eqnarray}

Similarly, on $\varepsilon_{11}$, we have
\begin{eqnarray}\label{thm4eq10}
	&&||\xi_{5M_{\beta}} (\hat{\beta}_1 - \beta_1)||_2 = O_p(s_\beta^{3/2}\log n) = o_p(\sqrt{s_\beta n}).
\end{eqnarray}

By (\ref{omega}), (\ref{term4}) and (\ref{thm4eq7})--(\ref{thm4eq10}), we have
\begin{eqnarray}\label{thm3eq19}
&&||B_{n\beta}^{-1}\omega_{M_{\beta}}||_2 \le \lambda_{\max}(B_{n\beta}^{-1}) ||\omega_{M_{\beta}}||_2= o(\frac{1}{n}\sqrt{s_\beta n})=o(\frac{\sqrt{s_\beta}}{\sqrt{n}}).
\end{eqnarray}
This along with (\ref{alpha1hat-alpha1}), (\ref{thm3eq8})--(\ref{thm3eq18}) and conditions in (\ref{x2x12infty}), (\ref{xx1a2norm2}) that
\begin{eqnarray*}
&&\displaystyle||\frac{1}{n\lambda_{2n}}\hat{Z}_2^T (Y - \hat{\Phi} - \hat{Z}_1 \hat{\beta}_1)||_\infty \le \frac{1}{n\lambda_{2n}}\{||X_{2\beta}^T W_\beta W(\hat{\theta}_1) X_{1\alpha}(\hat{\alpha}_1-\alpha_1)||_\infty\\
&+&||X_{2\beta}^T \Delta X_{1\beta} \{B_{n\beta}^{-1} nv_\beta+o(\sqrt{s_\beta}/\sqrt{n})\}||_\infty+O(n^{1-d_\beta}\sqrt{\log n}) + O(\sqrt{s_\beta n})\}\\
&\le&\frac{1}{n\lambda_{2n}}\{O(n^{1-d_\beta}\sqrt{\log n}) + O(\sqrt{s_\beta n})\}=o(1).
\end{eqnarray*}

Thus, (\ref{betaineqcon}) is verified. Finally by (\ref{term4}), (\ref{betasdcon}) holds. This concludes the proof.
\newpage
\section{Proof of Theorem 3} We only prove the asymptotic normality of $\hat{\beta}_1$ here. On the event $H_n$ defined in the proof of Theorem 2, it has been shown that $\hat{\beta}_1$ is a strict local minimizer of $\bar{Q}_n(\delta_\beta)$ and $\hat{\beta}_2 = 0$. It follows easily that $\nabla \bar{Q}_n(\hat{\beta}_1) = 0$. Thus, we have
\begin{eqnarray}\label{thm4eq0}
0 = \frac{1}{n} \hat{Z}_1^T (Y -\hat{\Phi}- \hat{Z}_1 \beta_1) - \frac{1}{n}\hat{Z}_1^T \hat{Z}_1 (\hat{\beta}_1 - \beta_1) - \lambda_{2n} \bar{\rho}_2(\hat{\beta}_1).
\end{eqnarray}

By (\ref{penalty3}), we have
\begin{eqnarray}\label{penalty}
||\lambda_{2n} \bar{\rho}_2(\hat{\beta}_1)||_2 \le \sqrt{s_\beta} \lambda_{2n} \bar{\rho}_2 (d_{n\beta}) = o_p(\frac{1}{\sqrt{n}}),
\end{eqnarray}
due to the monotonicity of the penalty function.

%

It follows from (\ref{omega}), (\ref{thm4eq7}), (\ref{thm4eq9}), (\ref{thm4eq10}) and that $\max(l_1,l_2)<\frac{1}{3}$.
\begin{eqnarray}\label{diffbeta}
||\omega_{M_{\beta}}||_2 = o_p(\sqrt{n}),
\end{eqnarray}

By (\ref{thm3eq5}) and (\ref{thm3eq6}), we have
\begin{eqnarray}\label{thm4eq11}
&&||Z_1^T (\hat{Z}_1 - Z_1) \beta_1||_2 = O_p(s_\alpha \sqrt{s_\beta n \log n}) = o_p(\sqrt{n}), \\ \label{thm4eq12}
&&||\eta_{2M_{\beta}}||_2=O(s_\alpha \sqrt{s_\beta})=o_p(\sqrt{n}).
\end{eqnarray}

We can further modify the right-hand order in $\varepsilon_{15}$ to be $o(\sqrt{n/s_\beta})$ since the corresponding condition (\ref{xphi2norm}) is strengthened, which gives
\begin{eqnarray}\label{thm4eq15}
||Z_1^T (\Phi-\hat{\Phi})||_2\le \sqrt{s_\beta} ||Z_1^T (\Phi-\hat{\Phi})||_\infty=o(\sqrt{n}).
\end{eqnarray}

Now it follows from (\ref{thm4eq0}), (\ref{penalty}), (\ref{diffbeta})--(\ref{thm4eq15}) that
\begin{eqnarray}\label{thm4eq14}
||\hat{Z}_1^T (Y - \hat{Z}_1 \beta_1) - \xi_{1M_{\beta}} - \xi_{2M_{\beta}} - \eta_{1M_{\beta}}||_2 = o_p(\sqrt{n}).
\end{eqnarray}

By (\ref{thm3eq11}), and that $\max(l_1, l_2)<\frac{1}{3}$, we have
\begin{eqnarray}\label{thm4eq13}
||\eta_{1M_{\beta}}+X_{1\beta}^T W(\hat{\theta}_1) \Delta X_{1\alpha} (\hat{\alpha}_1-\alpha_1)||_2=o(\sqrt{n}).
\end{eqnarray}

Combining (\ref{thm4eq14}), (\ref{thm4eq13}) together with
\begin{eqnarray*}
&\Mean||(\hat{Z}_1-Z_1)^T \epsilon||_2^2I_{\varepsilon} \le \displaystyle \sigma^2 \max_j \sqrt{s_\beta\lambda_{\max}\{X_{1\alpha}^T \diag(|x^j|) X_{1\alpha}\}} \\
&\sqrt{\lambda_{\max}\{X_{1\beta}^T \diag(|x^j|) X_{1\beta}\}} ||\hat{\alpha}_1-\alpha_1||_2=O(\sqrt{s_\alpha s_\beta n})=o(n),
\end{eqnarray*}
gives
\begin{eqnarray*}
B_{n\beta} (\hat{\beta}_1 -\beta_1) = Z_1^T \epsilon + Z_1^T (\mu-\Phi) - X_{1\beta}^T \diag(\mu-\hat{\Phi}) \Delta X_{1\alpha} (\hat{\alpha}_1-\alpha_1)+o_p(\sqrt{n}).
\end{eqnarray*}

The asymptotic normality of $\hat{\alpha}_1$ implies
\begin{eqnarray}\label{normalpha}
\label{asyalpha}B_{n\alpha} (\hat{\alpha}_1-\alpha_1) = X_{1\alpha}^T (A-\pi)+o_p(\sqrt{n}).
\end{eqnarray}

Let $B_{n\alpha\beta}=X_{1\beta}^T W \Delta X_{1\alpha}$, it follows from (\ref{cov3}) that
\begin{eqnarray}\label{lammax1}
&&\lambda_{\max}\{B_{n\beta}^{-1/2} B_{n\alpha\beta} B_{n\alpha}^{-1} B_{n\alpha\beta}^T B_{n\beta}^{-1/2}\}\\ \nonumber
&\le& \lambda_{\max}(\Delta^{\frac{1}{2}} X_{1\alpha} B_{n\alpha}^{-1} X_{1\alpha}^T \Delta^{\frac{1}{2}})\lambda_{\max}(B_{n\beta}^{-1/2}X_{1\beta}^T W^2 X_{1\beta}B_{n\beta}^{-1/2})\\ \nonumber
&=&\lambda_{\max}(B_{n\alpha}^{-1} B_{n\alpha})\lambda_{\max}(B_{n\beta}^{-1/2}X_{1\beta}^T W^2 X_{1\beta}B_{n\beta}^{-1/2})=O(1).
\end{eqnarray}

Combining (\ref{normalpha}), (\ref{lammax1}) with condition in (\ref{covphi3}) gives
\begin{eqnarray*}
B_{n\beta}^{1/2} (\hat{\beta}_1 - \beta_1) = B_{n\beta}^{-1/2} \{Z_1^T \epsilon + \xi_{2M_{\beta}} - B_{n\alpha\beta} B_{n\alpha}^{-1} X_{1\alpha}^T (A-\pi)\} + o_p(1).
\end{eqnarray*}

Therefore the covariance matrix of the above variable is $\Sigma_{22}$, which is $O(1)$ by (\ref{cov3}).

Define $\psi_{1i} = a^T A_{n2}^T B_{n\beta}^{-1/2} (A_i - \pi_i) x_{1\beta i} \epsilon_i$, $\psi_{2i} = a^T A_{n2}^T B_{n\beta}^{-1/2} (A_i - \pi_i) x_{1\beta i} (\mu_i -\Phi_i)$, $\psi_{3i}=a^T A_{n2}^T B_{n\beta}^{-1/2} B_{n\alpha\beta} B_{n\alpha}^{-1} (A_i - \pi_i) x_{1\alpha i}$. Now the asymptotic normality will follow if we can show $\sum_{i=1}^n \Mean|\psi_{ji}|^3=o(1)$ for $j=1,2,3$ by Lyapunov's theorem. Note that $\Mean|\varepsilon_i|^3 < \infty$ since the moment generating function of $\varepsilon_i$ exists, this together with (\ref{lyapunov3}) and (\ref{lyapunov3mu-phi}) gives
\begin{eqnarray*}
\displaystyle \sum_{i=1}^n \Mean|\psi_{1i}|^3 &\le& O(1) \sum_{i=1}^n ||a^T A_{n2}||_2^3 ||B_{n\beta}^{-1/2} x_{1\beta i}||_2^3 \\
&\le& O(1) \sum_{i=1}^n (x_{1\beta i}^T B_{n\beta}^{-1} x_{1\beta i})^{3/2} = o(1),\\
\displaystyle \sum_{i=1}^n \Mean|\psi_{2i}|^3 &\le& \sum_{i=1}^n ||a^T A_{n2}||_2^3 ||B_{n\beta}^{-1/2} x_{1\beta i} (\mu_i-\Phi_i)||_2^3 \\
&\le& O(1) \sum_{i=1}^n (x_{1\beta i}^T B_{n\beta}^{-1} x_{1\beta i})^{3/2}(\mu_i-\Phi_i)^3 = o(1).
\end{eqnarray*}

Besides, It follows from (\ref{lyapunov3}) and (\ref{lammax1}) that
\begin{eqnarray*}
&&\displaystyle \sum_{i=1}^n \Mean|\psi_{3i}|^3 \le \sum_{i=1}^n ||a^T A_{n2}||_2^3 ||B_{n\beta}^{-1/2} B_{n3} B_{n1}^{-1} x_{1\alpha i}||_2^3\\
&&\le O(1) \sum_{i=1}^n \{x_{1\alpha i}^T B_{n\alpha}^{-1} B_{n\alpha\beta}^T B_{n\beta}^{-1} B_{n\alpha\beta} B_{n\alpha}^{-1} x_{1\alpha i}\}^{3/2}\le O(1)\\
&&\sum_{i=1}^n (x_{1\alpha i}^T B_{n\alpha}^{-1} x_{1\alpha i})^{3/2} \lambda_{\max}\{B_{n\beta}^{-1/2}B_{n\alpha\beta}B_{n\alpha}^{-1}B_{n\alpha\beta}^T B_{n\beta}^{-1/2}\}^{3/2}=o(1).
\end{eqnarray*}

Thus, the Lyapunov condition is verified. Now it remains to show
\begin{eqnarray*}
\Var\left(\sum_{i=1}^n(\psi_{1i}+\psi_{2i}-\psi_{3i})\right)=\sigma^2 a^T A_{2n}^T A_{2n} a + a^T A_{2n}^T \Sigma_{22} A_{2n} a.
\end{eqnarray*}

Note that $\psi_{1i}$ is uncorrelated with $\psi_{2i}$ and $\psi_{3i}$, we have
\begin{eqnarray*}
&&\Var\left(\sum_{i=1}^n(\psi_{1i}+\psi_{2i}-\psi_{3i})\right)=\sum_{i=1}^n\Var(\psi_{1i})+\sum_{i=1}^n\Var(\psi_{2i}-\psi_{3i})\\
&&=\sum_{i=1}^n \Mean ||(A_i - \pi_i) x_{1\beta i}^T B_{n\beta}^{-1/2} A_{n2} a \varepsilon_i||_2^2\\
&&+\sum_{i=1}^n \Mean ||(A_i - \pi_i) [(\mu_i -\Phi_i) x_{1\beta i}^T-x_{1\alpha i}^T B_{n\alpha}^{-1} B_{n\alpha\beta}^T ] B_{n\beta}^{-1/2} A_{n2} a||_2^2\\
&&=a^T A_{n2}^T B_{n\beta}^{-1/2} X_{1\beta}^T W (\Delta-\Delta X_{1\alpha} B_{n\alpha}^{-1} X_{1\alpha}^T \Delta) W X_{1\beta} B_{n\beta}^{-1/2} A_{n2} a\\
&&+\sigma^2 a^T A_{n2}^T A_{n2} a=\sigma^2 a^T A_{2n}^T A_{2n} a + a^T A_{2n}^T \Sigma_{22} A_{2n} a.
\end{eqnarray*}

This completes the proof.
\newpage
\section{Proof of Theorem 4}
Similar to the prove of Theorem 3, under the given conditions, with probability goes to $1$, we have
\begin{eqnarray}
\label{betaest}\hat{\beta}_2=0, \quad ||\hat{\beta}_1-\beta_1||_2=O(\frac{\sqrt{s_\beta}}{\sqrt{n}}).
\end{eqnarray}

In addition, under the conditions in Theorem 3, $B_{n\beta}^{-1/2} (\hat{\beta}_1-\beta_1)$ is asymptotically equivalent to
\begin{eqnarray}\label{thm5eq1}
B_{n\beta}^{-1/2} \{Z_1^T \epsilon + \xi_{2M_{\beta}} - B_{n\alpha\beta} B_{n\alpha}^{-1} X_{1\alpha}^T (A-\pi)\}.
\end{eqnarray}

By the definition of $\hat{V}_n$, we have
\begin{eqnarray}\label{thm5eq2}
~~\sqrt{n}[\hat{V}_n-V_n(\beta_0)]&=&\frac{1}{\sqrt{n}}\sum_i [e_i+x_i^T (\hat{\beta}-\beta_0)(I(x_i^T \beta>0)-A_i)]\\ \label{thm5eq3}
&+&\frac{1}{\sqrt{n}}\sum_i x_i^T \hat{\beta} [I(x_i^T \hat{\beta}>0)-I(x_i^T \beta_0>0)].
\end{eqnarray}

We break the proof into two steps. In the first step, we show (\ref{thm5eq3}) is $o_p(1)$. Next, we establish the asymptotic normality of the right-hand side of (\ref{thm5eq2}).
\quad\newline

\textit{Step 1}. Note that $(\ref{thm5eq3})$ is always nonnegative, it suffices to provide an upper bound. Since $\sum_i x_i^T \beta_0 [I(x_i^T \hat{\beta}>0)-I(x_i^T \beta_0>0)]\le 0$, (\ref{thm5eq3}) is smaller than
\begin{eqnarray}\label{thm5eq4}
\sum_i [x_i^T \hat{\beta}-x_i^T \beta_0] [I(x_i^T \hat{\beta}>0)-I(x_i^T \beta_0>0)].
\end{eqnarray}

We further partition (\ref{thm5eq4}) into the following two pieces.
\begin{eqnarray*}
I_1=\sum_i [x_i^T \hat{\beta}-x_i^T \beta_0] [I(x_i^T \hat{\beta}>0)-I(x_i^T \beta_0>0)]I(x_i^T \beta_0\le n^{-1/4}),\\
I_2=\sum_i [x_i^T \hat{\beta}-x_i^T \beta_0] [I(x_i^T \hat{\beta}>0)-I(x_i^T \beta_0>0)]I(x_i^T \beta_0> n^{-1/4}).
\end{eqnarray*}

It follows from Cauchy-Swartz inequality that $I_1$ is smaller than
\begin{eqnarray*}
&&\sqrt{\sum_i (x_i^T \hat{\beta}-x_i^T \beta_0)^2}\sqrt{\sum_i I(x_i^T \beta_0\le n^{-1/4})^2}\\
&\le& \sqrt{(\hat{\beta}_1-\beta_1)^T X_{1\beta}^T X_{1\beta}(\hat{\beta}_1-\beta_1)} \sqrt{\sum_i I(x_i^T \beta_0\le n^{-1/4})}\\
&=& O_p(\sqrt{s_\beta}) C^\prime n^{-1/4}=o_p(1),
\end{eqnarray*}
followed by (\ref{betaest}), Condition 7 and that $s_\beta=o(n^{1/4})$. As for $I_2$, note that for arbitrary real numbers $a$, $b$, $|a-b|>|a|$ if $ab < 0$. An elementary calculation shows it is smaller than
\begin{eqnarray*}
&&\frac{1}{\sqrt{n}}\sum_{i=1}^n |x_i^T (\hat{\beta}-\beta_0)| I(|x_i^T (\hat{\beta}-\beta_0)| > |x_i^T \beta_0|) I(|x_i^T \beta_0|>n^{-1/4})\\
&\le& \frac{1}{\sqrt{n}}\sum_{i=1}^n [x_i^T (\hat{\beta}-\beta_0)]^2 / |x_i^T \beta_0| I(|x_i^T \beta_0|>n^{-1/4})\\
&\le&=\frac{1}{n^{1/4}} [x_i^T (\hat{\beta}-\beta_0)]^2=O_p(s_\beta/n^{1/4})=o_p(1),
\end{eqnarray*}
where the inequality in the second line follows from the fact that $I(|x_i^T (\hat{\beta}-\beta_0)| > |x_i^T \beta_0|)\le |x_i^T (\hat{\beta}-\beta_0)|/|x_i^T \beta_0|$. This shows (\ref{thm5eq3}) is $o_p(1)$.
\quad\newline

\textit{Step 2}. Using similar arguments in the proof of Lemma 2, we can show with probability at least $1-s_\beta/n$, the following event holds:
\begin{eqnarray}\label{thm5eq5}
\max_{j=1,\dots,s_\beta} |(A-\pi)^T x^j|=O(\sqrt{n\log n}).
\end{eqnarray}

On the events (\ref{betaest}) and (\ref{thm5eq5}), we have
\begin{eqnarray*}
&\displaystyle \frac{1}{\sqrt{n}}\sum_i [x_i^T (\hat{\beta}-\beta_0)(I(x_i^T \beta>0)-A_i)- x_i^T (\hat{\beta}-\beta_0)(I(x_i^T \beta>0)-\pi_i)]\\
&\displaystyle \le ||X_{1\beta}^T (A-\pi)||_2 ||\hat{\beta}_1-\beta_1||_2\le O(\sqrt{s_\beta}) \max_{j=1,\dots,s_\beta} |(A-\pi)^T x^j| O(\frac{\sqrt{s_\beta}}{\sqrt{n}}),
\end{eqnarray*}
which is $o(1)$. Therefore, it suffices to establish the asymptotic normality for
\begin{eqnarray*}
\frac{1}{\sqrt{n}}\sum_i [e_i+x_i^T (\hat{\beta}-\beta_0)(I(x_i^T \beta>0)-\pi_i)],
\end{eqnarray*}
or equivalently,
\begin{eqnarray}\label{thm5eq6}
\frac{1}{\sqrt{n}}\sum_i [e_i+v_n^T X_{1\beta} (\hat{\beta}-\beta_0)].
\end{eqnarray}

Under the condition (\ref{lamminx}) and $\lambda_{\max}(X_{1\beta}^T X_{1\beta})=O(n)$, we have
\begin{eqnarray*}
||B_{n\beta}^{-1/2} X_{1\beta}^T v_n||_2\le \sqrt{\lambda_{\max}(X_{1\beta}^T X_{1\beta})/\lambda_{\min}(B_{n\beta})}||v_n||_2=O(1),
\end{eqnarray*}
which together with (\ref{thm5eq1}) implies that (\ref{thm5eq6}) is asymptotically equivalent to
\begin{eqnarray}\label{thmeq7}
\frac{1}{\sqrt{n}}\sum_i e_i +v_n^T X_{1\beta} B_{n\beta}^{-1} [Z_1^T e + \xi_{2M_{\beta}} - B_{n\alpha\beta} B_{n\alpha}^{-1} X_{1\alpha}^T (A-\pi)].
\end{eqnarray}

Using similar arguments in the proof of Theorem 3, we can show the Lyaponuv's condition for (\ref{thmeq7}) holds. By the independence between $e$ and $A-\pi$, the variance of (\ref{thmeq7}) is equal to $v_1^2+v_2^2$, where
\begin{eqnarray*}
v_1^2&=&\Var\left(\frac{1}{\sqrt{n}}\sum_i e_i +v_n^T X_{1\beta} B_{n\beta}^{-1} Z_1^T e\right),\\
v_2^2&=&\Var\left(v_n^T X_{1\beta} B_{n\beta}^{-1}[\xi_{2M_{\beta}} - B_{n\alpha\beta} B_{n\alpha}^{-1} X_{1\alpha}^T (A-\pi)]\right).
\end{eqnarray*}

Using the chain rule for conditional expectation, $v_1^2$ reduces to
\begin{eqnarray*}
&&\Mean\left\{\Var\left(\frac{1}{\sqrt{n}}\sum_i e_i +v_n^T X_{1\beta} B_{n\beta}^{-1} Z_1^T e\right)|Z_1\right\}\\
&=&\sigma^2+\Mean (v_n^T X_{1\beta} B_{n\beta}^{-1} Z_1^T Z_1 B_{n\beta}^{-1} X_{1\beta}^T v_n)+\frac{1}{\sqrt{n}}\Mean(v_n^T X_{1\beta} B_{n\beta}^{-1} Z_1^T \mathbf{1})\\
&=&\sigma^2+v_n^T X_{1\beta} B_{n\beta}^{-1}X_{1\beta}^T v_n,
\end{eqnarray*}
while $v_2^2$ is equal to
\begin{eqnarray*}
&&v_n^T X_{1\beta} B_{n\beta}^{-1/2}\Var (B_{n\beta}^{-1/2}[\xi_{2M_{\beta}} - B_{n\alpha\beta} B_{n\alpha}^{-1} X_{1\alpha}^T (A-\pi)])B_{n\beta}^{-1/2} X_{1\beta}^T v_n\\
&=&v_n^T X_{1\beta} B_{n\beta}^{-1/2}\Sigma_{22} B_{n\beta}^{-1/2} X_{1\beta}^T v_n,
\end{eqnarray*}
by the definition of $\Sigma_{22}$. This completes the proof.

\newpage
\section{Proof of Lemma 2}
Define
\begin{eqnarray*}
	\varepsilon_1 &=& \left\{||X_{1\alpha}^T (A - \pi)||_\infty \le \frac{1}{\sqrt{2}}\sqrt{n\log n} \right\},\\
	\varepsilon_2 &=& \left\{||X_{2\alpha}^T (A - \pi)||_\infty \le \frac{1}{\sqrt{2}}n^{1-d_\beta}(\log n)^\frac{1}{2}\right\},
\end{eqnarray*}
similar arguments in the proof of Theorem 2 in \cite{fan2011} gives $\prob(\varepsilon_1 \cup \varepsilon_2|X) \ge 1-c_0/(n+p)$ for some $c_0>0$. Conditional on $\varepsilon_1 \cup \varepsilon_2$, $\hat{\alpha}$ is shown to have weak oracle property.

Consider events
\begin{eqnarray*}
	&&\varepsilon_3 = \left\{||\xi_{1M_{\beta}}||_\infty = O(\sqrt{n \log n})\right\},
	\varepsilon_4 = \left\{||\xi_{1M_{\beta}^c}||_\infty = O(n^{1-d_\beta} \sqrt{\log n})\right\},\\
	&&\varepsilon_5 = \left\{||\xi_{2M_{\beta}}||_\infty = O(\sqrt{n \log n})\right\},
	\varepsilon_6 = \left\{||\xi_{2M_{\beta}^c}||_\infty = O(n^{1-d_\beta} \sqrt{\log n})\right\},\\
	&&\varepsilon_7 = \left\{||\xi_{3M_{\theta}^\prime}||_\infty = O(\sqrt{n \log n})\right\},
	\varepsilon_8 = \left\{||\xi_{3M_{\theta}^{c\prime}}||_\infty = O(n^{1-d_\theta} \sqrt{\log n})\right\},\\
	&&\varepsilon_{9} = \left\{||\xi_{4M_{\beta}}||_\infty = O(\frac{n^{1/2+\gamma_\alpha}}{\sqrt{\log n}})\right\},
	\varepsilon_{10} = \left\{||\xi_{4M_{\beta}^c}||_\infty = O(\frac{n^{1-d_\beta+\gamma_\alpha}}{\sqrt{\log n}})\right\},\\
	&&\varepsilon_{11} = \left\{||\xi_{5M_{\beta}}||_\infty = O(\frac{n^{1/2+\gamma_\beta}}{\sqrt{\log n}})\right\},
	\varepsilon_{12} = \left\{||\xi_{5M_{\beta}^c}||_\infty = O(\frac{n^{1-d_\beta+\gamma_\beta}}{\sqrt{\log n}}) \right\},\\
	&&\varepsilon_{13} = \left\{\sup_{\delta \in H_\theta}||\xi_{7M_{\beta}}(\delta)||_\infty = O(\sqrt{n \log n})\right\},\\
	&&\varepsilon_{14} = \left\{\sup_{\delta \in H_\theta}||\xi_{7M_{\beta}^c}(\delta)||_\infty = O(n^{1-d_\beta} \sqrt{\log n})\right\},\\
	&&\varepsilon_{15} = \left\{\sup_{\delta \in H_\theta}||\xi_{8M_{\theta}^\prime}(\delta)||_\infty = O(\sqrt{n \log n})\right\},\\
	&&\varepsilon_{16} = \left\{\sup_{\delta \in H_\theta}||\xi_{8{M_{\theta}^\prime}^c(\delta)}||_\infty = O(n^{1-d_\beta} \sqrt{\log n})\right\}.
\end{eqnarray*}
Denote $\omega=\max_i ||e_i||_{\psi_1}$. Since $||(A_i-\hat{\pi}_i) x_i||_2\le ||x_i||_2=\sqrt{n}$, $||(A_i-\hat{\pi}_i) x_i||_\infty\le ||x_i||_\infty$, take $\epsilon=4\sqrt{n\log n}$, it follows from Lemma 1 and (\ref{maxentry})
that
\begin{eqnarray}\label{lemma2eq1}
\prob(||\xi_{1M_\beta}||_\infty>\epsilon)\le 2s_\beta \max_{j\in M_\beta} \prob(|\xi_{1}^j|>\epsilon)\\\nonumber
\le 2s_\beta\exp\left(\frac{16\omega^2n\log n}{4n\omega^2+O(n)}\right)\le \frac{2s_\beta}{n^2}\le\frac{2}{n}.
\end{eqnarray}

Using similar arguments we can show $\prob(\varepsilon_4)\ge 1-2/p$, which together with (\ref{lemma2eq1}) entails
\begin{eqnarray*}
	\prob(\varepsilon_3\cup\varepsilon_4)\ge 1-\frac{2}{n+p}.
\end{eqnarray*}

Similarly, it follows from (\ref{xmu2norm}) that
$$\prob(\varepsilon_5 \cup \varepsilon_6) \ge 1-\frac{2}{n+p}.$$

Besides, by (\ref{phi2norm}) and (\ref{phimax}),
$$\prob(\varepsilon_7 \cup \varepsilon_8) \ge 1-\frac{2}{n+p}.$$

Now we show $\varepsilon_9\cup \varepsilon_{10}\cup \varepsilon_{11}\cup \varepsilon_{12}$ hold with large probability. It follows by the definition of $\xi_4$ that
\begin{eqnarray*}
\max_{j\in M_\alpha}||(||\xi^j_{4M_\beta}||_1)||_{\psi_2}\le \max_j \sum_{k \in M_\beta} ||\xi_{4k}^j||_{\psi_2}=\max_{j}\sum_{k \in M_\beta}||x^k \circ x^j \circ (X\beta_0)||_2.
\end{eqnarray*}

Take $t=\max_{j}\sum_{k \in M_\beta}||x^k \circ x^j \circ (X\beta_0)||_2$, it follows by the definition of $||\cdot||_{\psi_2}$ that
\begin{eqnarray*}
\prob(||\xi^j_{4M_\beta}||_1\le \sqrt{2\log n}t)\le \frac{\exp([||\xi^j_{4M_\beta}||_1/t]^2)}{[\sqrt{2\log n}t/t]^2}\le \frac{2}{n^2},
\end{eqnarray*}
which together with Bonferroni's inequality and Condition (\ref{xxaxb2norm}) entails
\begin{eqnarray*}
\prob(\varepsilon_9)\ge 1-\sum_{j \in M_\alpha} \prob(||\xi^j_{4M_\beta}||_1\le \sqrt{2\log n}t)\ge 1-\frac{2}{n}.
\end{eqnarray*}
By the same argument and condition (\ref{xx2norm}), we have $\prob(\varepsilon_{10}\cup \varepsilon_{11}\cup \varepsilon_{12}) \ge 1-6/(n+p)$. Taking the hypercube in Proposition \ref{propdeviation} to be $H_\theta$, $z_i=e_i$, $A(h)=\phi_1(\delta)$, $\delta \in H_\theta$, $t=\sqrt{n\log n}$, clearly, by condition (\ref{xphi2norm}), for sufficiently large $n$,
\begin{eqnarray*}
t\ge 3\sigma^2 n^{-\gamma_\theta}\log n\max_j\sum_{l\in M_\theta} \sup_{\delta \in H_\theta}||x^j \circ \phi^l(\delta)||_2.
\end{eqnarray*}

It follows by Proposition \ref{propdeviation} that for some sufficiently large constant $M$,
\begin{eqnarray*}
	\prob(\varepsilon_{13})\ge \prob(\sup_{\delta \in H_\theta}||\xi_{7M_{\beta}}(\delta)\le Mt)\ge 1-\frac{2}{n}-n\exp\left(\frac{2M^2t^2}{Mn^2}\right)\ge 1-\frac{4}{n}.
\end{eqnarray*}
Using the same argument, under condition (\ref{dphi2norm}) and that $\log p = O(n^{1-2d_\beta})$, $\log q = O(n^{1-2d_\theta})$, we have $\prob(\varepsilon_{14}\cup \varepsilon_{15} \cup \varepsilon_{16})\to 1-8/(n+q)$. This completes the proof.
\newpage

\newpage

\bibliographystyle{biom}
\bibliography{2}
\end{document}